\documentclass{article}
\usepackage[margin=1in]{geometry}
\usepackage{amsmath,amssymb}
\usepackage{graphicx}
\usepackage{booktabs}
\usepackage{tabularx}
\usepackage{natbib}
\usepackage{tikz}
\usepackage{placeins}
\usepackage{bbm}
\definecolor{flowblue}{HTML}{2A78D6}
\definecolor{floworange}{HTML}{E96B39}
\definecolor{flowgreen}{HTML}{1A9B70}
\definecolor{flowink}{HTML}{252525}
\definecolor{flowgray}{HTML}{F5F6F7}
\usetikzlibrary{positioning, arrows.meta}
\usepackage[colorlinks=true,citecolor=blue,linkcolor=blue,urlcolor=blue]{hyperref}
\usepackage[nameinlink,capitalize]{cleveref}
\newcommand{\Dcf}{\widehat{D}_{\mathrm{CF5}}}

\title{When Does Dynamic Ensembling Pay Off? Diagnosing Regionwise Gains in Regression under Distribution Shift}
\author{Tianxin Zhou \and Ruixi Lin}
\date{August 2026}

\begin{document}

\maketitle

\begin{abstract}
Whether input-dependent (``dynamic'') combination of a regression model pool beats the best static blend depends on the shift and is rarely known before deployment. Can a small labeled target-domain probe tell us when reallocating trust across regions of the input space will pay off? We answer this with $\Dcf$, which estimates from the probe the cross-fitted gain of the regionwise convex combination over the best static convex blend: the realizable value of deciding, region by region, whom to trust. 

Across a frozen suite of 12 dataset-shift pairs spanning spatial, temporal, domain and feature-cluster shifts (five used for development and seven reserved in two preregistered held-out batches), $\Dcf$ predicts realized regionwise test gains with dataset-level Spearman $+0.98$ (bootstrap 95\% CI $[+0.83, +1.00]$; permutation $p=5\times10^{-5}$; calibration slope $0.94$), including two cases that overturned our preregistered expectations. The relationship remains strong in a 16-pair sensitivity analysis that adds four pairs from the prospective selector batch (Spearman $+0.83$). Alternative probe diagnostics reach at most $+0.66$. The contrast isolates regional trust reallocation: its correlation is $+0.98$ for regionwise-convex gain but $+0.01$ for the incremental value of smooth covariate-dependent stacking after affine correction.

A controlled generator shows that dynamic gains arise from the interaction of shift heterogeneity and local competence and increase with shift severity; in the tested grid, the transition occurs between 128 and 256 probe labels. Complementing the diagnostic, the Probe-Validated Ensemble Selector chooses among a static affine stacker and dynamic realizers, deploying a candidate only when a held-out lower confidence bound clears the static-convex floor. In a preregistered prospective batch, it matched or improved that floor in all 12 runs; the two deployments reduced test risk by $11\%$ and $16\%$, while the gate rejected a candidate whose unconditional deployment incurred over 30 times the static loss.

We release OpenRegShift, a reproducible evaluation harness for regression ensembles under distribution shift. 

\end{abstract}

\section{Introduction}
\label{sec:intro}
A practitioner has a pool of regression models trained on source-domain data and a small labeled sample from a shifted target domain. Should the models be averaged, combined with one global set of weights, or trusted differently in different parts of the input space? Input-dependent combination is appealing: if different models stay accurate in different regions, a dynamic combiner can exploit their local complementarity. The same flexibility can amplify probe noise when no such structure exists. The practical question is therefore when the target data support dynamic combination.

This question matters because static ensembling is a strong baseline. Uniform averaging and deep ensembles are accurate and robust \citep{lakshminarayanan2017simple,ovadia2019can}, while stacking learns a single combination from validation data \citep{wolpert1992stacked}. Dynamic alternatives such as mixture-of-experts, dynamic regressor selection, and covariate-dependent stacking are far more expressive \citep{jacobs1991adaptive,moura2019evaluating,wakayama2025ensemble}. Yet their gains vary sharply across shifts, and neither shift severity nor model disagreement reveals whether the local structure they need is present. 

We formulate this deployment decision as an estimable comparison. From a labeled probe, we learn a partition of the input space and compare the best static convex blend with a regionwise convex blend, with both fitted and evaluated out of fold. The resulting cross-fitted risk difference, $\Dcf$, estimates the realizable value of deciding, region by region, whom to trust. Its sign answers the decision for this regional mechanism: a positive value means that local trust reallocation survives the estimation cost of a finite probe.

Across a frozen suite of 12 dataset/shift pairs spanning spatial, temporal, domain and feature-cluster shifts, $\Dcf$ predicts realized regionwise gains with dataset-level Spearman correlation $+0.98$, including two preregistered held-out shifts that overturned our recorded expectations. The correlation remains $+0.83$ in a sensitivity analysis that adds four pairs from the prospective selector batch. Simpler probe summaries reach at most $+0.66$. The diagnostic is also mechanism-specific: it correlates at $+0.98$ with the regionwise gain it measures and at $+0.01$ with the dynamic increment of covariate-dependent stacking beyond affine correction. It identifies the value of reallocating trust across regions, while deployment validation determines which available mechanism should realize the gain.

A controlled generator explains when this value exists. Gains arise from the interaction of shift heterogeneity and local competence; varying either condition alone produces little. Gains increase with shift severity, with the transition occurring between 128 and 256 probe labels in the tested grid. 

We complement the diagnosis with a deployment procedure. The Probe-Validated Ensemble Selector evaluates a static affine stacker and several dynamic realizers against a static convex floor. It deploys a candidate only when its held-out lower confidence bound clears that floor. In a preregistered prospective batch, the selector held the floor in all 12 runs. Both dynamic deployments realized their validated gains, improving over the floor by $11\%$ and $16\%$, while the gate rejected a candidate whose unconditional loss exceeded thirty times the static loss.

Our contributions are:
\begin{itemize}
\item $\Dcf$, a cross-fitted diagnostic of the realizable regionwise gain over static convex blending, validated across diverse regression shifts.
\item The Probe-Validated Ensemble Selector, which converts a labeled target probe into a validated choice among a static floor and complementary dynamic realizers.
\item Controlled factorial and dose-response experiments that isolate the conjunctive mechanism behind dynamic gains and measure how realizability changes with shift severity and probe budget.
\item OpenRegShift, a reproducible evaluation harness covering 16 dataset/shift pairs, a broad baseline suite, preregistered held-out evaluations, and one-command result verification. 
\end{itemize}

\section{The diagnostic}
\label{sec:diagnostic}

Let a frozen pool of $K$ source-trained regression models produce the prediction vector
\begin{equation}
\mu(x)=\bigl(\mu_1(x),\ldots,\mu_K(x)\bigr)^\top
\in \mathbb{R}^K.
\end{equation}

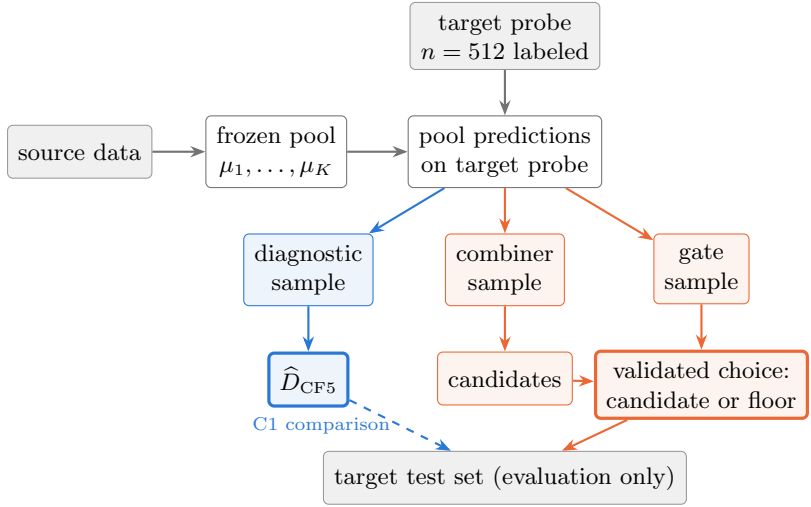
\begin{figure}[t]
\centering
\definecolor{diagblue}{HTML}{2A78D6}
\definecolor{selorange}{HTML}{EB6834}
\begin{tikzpicture}[
  font=\small,
  nodebox/.style={draw=black!55, rounded corners=2pt, align=center,
                  inner sep=4pt, minimum height=7mm},
  nodedata/.style={nodebox, fill=black!6, draw=black!40},
  diagbox/.style={nodebox, draw=diagblue, fill=diagblue!8},
  diagres/.style={diagbox, very thick},
  selbox/.style={nodebox, draw=selorange, fill=selorange!8},
  selres/.style={selbox, very thick},
  arr/.style={-{Stealth[length=2mm]}, thick, black!55},
  darr/.style={arr, diagblue},
  sarr/.style={arr, selorange}]

\node[nodedata] (src) {source data};
\node[nodebox, right=7mm of src] (pool) {frozen pool\\$\mu_1,\ldots,\mu_K$};
\node[nodebox, right=8mm of pool] (pred) {pool predictions\\on target probe};
\node[nodedata, above=6mm of pred] (probe) {target probe\\$n=512$ labeled};

\node[diagbox, below=6mm of pred, xshift=-26mm] (pd) {diagnostic\\sample};
\node[selbox, below=6mm of pred] (fit) {combiner\\sample};
\node[selbox, below=6mm of pred, xshift=26mm] (gate) {gate\\sample};

\node[diagres, below=6mm of pd] (d) {$\Dcf$};
\node[selbox, below=6mm of fit] (cand) {candidates};
\node[selres, below=6mm of gate] (dec) {validated choice:\\candidate or floor};

\node[nodedata, below=6mm of cand] (test) {target test set (evaluation only)};

\draw[arr] (src) -- (pool);
\draw[arr] (pool) -- (pred);
\draw[arr] (probe) -- (pred);
\draw[darr] (pred) -- (pd);
\draw[sarr] (pred) -- (fit);
\draw[sarr] (pred) -- (gate);
\draw[darr] (pd) -- (d);
\draw[sarr] (fit) -- (cand);
\draw[sarr] (cand) -- (dec);
\draw[sarr] (gate) -- (dec);
\draw[darr, dashed] (d) -- 
node[left, font=\scriptsize] {C1 comparison} (test);
\draw[sarr] (dec) -- (test);
\end{tikzpicture}
\caption{Data flow. The pool is trained on source data only and is evaluated at the target probe points. The probe is split once into a diagnostic sample (blue), a combiner sample and a gate sample (orange). The diagnostic branch yields $\Dcf$; the selector branch fits candidates and validates them against the static convex floor. The test set is not used for fitting or selection. The two branches are complementary modules: the selector does not take $\Dcf$ as input. The dashed blue arrow denotes the final C1 characterization analysis; test observations do not enter the diagnostic.}
\label{fig:pipeline}
\end{figure}

At deployment, we observe a labeled probe from the target domain of total size $n=512$. The probe is split once into a diagnostic sample, a combiner sample, and a gate sample. The diagnostic sample supports the analysis in this section; the other two are reserved for the ensemble selector in \cref{sec:selector}. A disjoint target test set is used only for final evaluation. We use squared error on standardized targets. \Cref{fig:pipeline} summarizes the data flow, and the complete access and splitting protocol appears in \cref{tab:protocol}.

\paragraph{Population headroom.}
Let
\begin{equation}
\Delta_K=\left\{w\in\mathbb{R}^K:
w_k\geq 0,\;
\sum_{k=1}^K w_k=1
\right\}
\end{equation}
denote the probability simplex, and let
$\pi:\mathcal{X}\rightarrow\{1,\ldots,J\}$ be a partition of the target input space. We set $J=8$ in all primary experiments and report a sweep over $J$ in \cref{app:jsweep}. The optimal static convex risk is
\begin{equation}
R_{\mathrm{static}}^\star=\min_{w\in\Delta_K}
\mathbb{E}_T
\left[
\bigl(Y-w^\top\mu(X)\bigr)^2
\right].
\end{equation}
The corresponding regionwise convex risk is
\begin{equation}
R_{\mathrm{regional}}^\star(\pi)=\min_{w_1,\ldots,w_J\in\Delta_K}\mathbb{E}_T
\left[
\bigl(Y-w_{\pi(X)}^\top\mu(X)\bigr)^2
\right].
\end{equation}
Because the regionwise class contains every static convex blend,
\begin{equation}
D(\pi)=R_{\mathrm{static}}^\star-R_{\mathrm{regional}}^\star(\pi)
\geq 0.
\end{equation}
We call $D(\pi)$ the \emph{population regionwise headroom}. It measures the structural value of reallocating model trust across the regions defined by $\pi$.

\paragraph{Realizable gain at probe budget $m$.}
Population headroom does not account for finite-sample estimation. Let $\widehat{\pi}_n$ denote the partition learned from the covariates of all $n=512$ target-probe observations; partition fitting uses no target labels and no target test observations. Let $S_m=\mathcal{P}_D$ denote the diagnostic sample, where $m=205$. Let $\mathcal{A}_{\mathrm{static}}(S_m)$ denote static convex fitting, and let $\mathcal{A}_{\mathrm{regional}}(S_m,\widehat{\pi}_n)$ denote regionwise convex fitting under the learned partition. We set the minimum regional sample size to $m_{\min}=5$. Any region containing fewer than $m_{\min}$ observations from $S_m$ reuses the global static weights. We define
\begin{equation}
G_m(\widehat{\pi}_n)=\mathbb{E}
\left[
    R_T\!\left(\mathcal{A}_{\mathrm{static}}(S_m)\right)
    -
    R_T\!\left(\mathcal{A}_{\mathrm{regional}}(S_m,\widehat{\pi}_n)\right)
    \;\middle|\;
    \widehat{\pi}_n
\right].
\end{equation}
The expectation is over the labeled diagnostic sample and its allocation, conditional on the learned partition. Unlike population headroom, $G_m(\widehat{\pi}_n)$ can be negative: the regional advantage must be large enough to offset weight estimation. Its sign therefore answers the deployment question for the regional mechanism. The quality of $\widehat{\pi}_n$ enters through the conditional gain, and the cost of learning the partition is isolated separately in
\cref{sec:mechanism}.

\paragraph{Cross-fitted estimation.}
We estimate the finite-budget gain using three repetitions of region-stratified five-fold cross-fitting within the diagnostic sample. The partition $\widehat{\pi}_n$ is learned once from all probe covariates and held fixed throughout. Partition fitting uses no target labels. Within every repetition and fold, the static and regional weights are fitted without the fold labels and evaluated on that fold.

For repetition $r\in\{1,2,3\}$, let $\{\mathcal{I}_{rq}\}_{q=1}^{5}$ denote the five folds. Let $\widehat f_{\mathrm{static}}^{(-r,q)}$ and $\widehat f_{\mathrm{regional}}^{(-r,q)}$ denote the two combiners fitted without fold $\mathcal{I}_{rq}$. The diagnostic is
\begin{equation}
\Dcf=\frac{1}{15}
\sum_{r=1}^{3}
\sum_{q=1}^{5}
\frac{1}{|\mathcal{I}_{rq}|}
\sum_{i\in\mathcal{I}_{rq}}
\left[
\ell\!\left(
y_i,
\widehat f_{\mathrm{static}}^{(-r,q)}(x_i)
\right)-\ell\!\left(y_i,\widehat f_{\mathrm{regional}}^{(-r,q)}(x_i)
\right)
\right].
\end{equation}
Here, $\ell(y,\widehat y)=(y-\widehat y)^2$. A positive value favors regionwise combination, while a value at or below zero favors the static convex blend.

Each fold model is trained on approximately $4m/5$ diagnostic observations. Thus, $\Dcf$ estimates the same regional contrast at a slightly smaller training budget than the final refit on all $m$ observations.

\paragraph{Why cross-fitting matters.}
An in-sample contrast rewards additional regional flexibility mechanically, even when that flexibility does not generalize. Cross-validation separates weight fitting from loss evaluation and limits the optimism created by evaluating a flexible combiner on the training labels \citep{arlot2010survey}. We compare the two estimators in \cref{sec:c1}. 

We define the effective number of estimable regions as
\begin{equation}
J_{\mathrm{eff}}=\sum_{j=1}^{J}
\mathbf{1}\!\left\{
\sum_{i\in\mathcal{P}_D}
\mathbf{1}\!\left\{
\widehat{\pi}_n(x_i)=j
\right\}
\geq m_{\min}
\right\}.
\end{equation}
When $J_{\mathrm{eff}}\leq 1$, the regional mechanism cannot be diagnosed, so $\Dcf$ is reported as non-estimable rather than assigned a misleading value of zero. When $J_{\mathrm{eff}}<J/2$, we retain the estimate but flag the partition as having reduced resolution. Under the setting $J=8$, the reduced-resolution flag is therefore triggered when fewer than four regions are estimable.

\paragraph{Alignment with realized gain.}
For the characterization experiments, \emph{realized regional gain} is defined as
\begin{equation}
\widehat G_{\mathrm{test}}=\widehat R_{\mathrm{test}}
\left(
\widehat f_{\mathrm{static}}
\right)-\widehat R_{\mathrm{test}}
\left(\widehat f_{\mathrm{regional}}
\right).
\end{equation}
Both combiners are refitted on the complete diagnostic sample and evaluated on the disjoint target test set. Thus, $\Dcf$ and realized gain compare the same static and regional classes under the same partition and loss. Their systematic budget difference is that cross-fitting trains each fold model on approximately $4m/5$ observations, whereas the reported test gain uses all $m$ diagnostic observations.

\section{Setup and protocol}
\label{sec:setup}
We study regression under distribution shift with a fixed pool of source-trained models and a limited labeled target probe. For each dataset/shift pair, let $P_S$ and $P_T$ denote the source and target distributions. The direction of the shift is defined before model training using time, geography, monitoring site, subject identity, domain identity, or a feature-cluster holdout. 

Source observations are divided into training and validation sets. All base regressors and preprocessing transformations are fitted using source data and frozen before target labels are accessed. Target labels may be used to diagnose and fit ensemble combinations, but never to update the base regressors. This protocol isolates the value of combining an existing model pool under shift.
The primary real-data experiments use a heterogeneous pool of five regressors: gradient boosting, random forest, ridge regression, a multilayer perceptron, and nearest neighbors. Each dataset/shift pair is evaluated with seeds $\{0,1,2\}$, and losses are measured on targets standardized using source-training statistics. Dataset construction and model-pool specifications appear in \cref{app:datasets,app:model-pool}.

\subsection{Target probe and access protocol}
\label{sec:access}

At deployment, exactly $n=512$ labeled target observations are assigned to the probe. A seeded permutation divides the probe into three disjoint samples:
\begin{equation}
|\mathcal{P}_D|=205,\qquad
|\mathcal{P}_C|=215,\qquad
|\mathcal{P}_G|=92.
\end{equation}

The diagnostic sample $\mathcal{P}_D$ receives 40\% of the probe. The remaining 60\% forms the method sample, which is divided 70/30 into the combiner sample $\mathcal{P}_C$ and gate sample $\mathcal{P}_G$. All remaining target observations form the test set. We fix the absolute probe budget across datasets rather than fixing its fraction of the available target data.

The regional partition is fitted by $k$-means with $J=8$ using the covariates of all 512 probe observations. Inputs are standardized using source-training statistics. Partition fitting uses no target labels and no target test observations. The learned partition is then held fixed for the diagnostic and selector branches.

\begin{table}[t]
\centering
\footnotesize
\caption{Frozen information-access protocol. Target test observations are unavailable until final evaluation.}
\begin{tabularx}{\linewidth}{@{}p{0.22\linewidth}p{0.22\linewidth}X@{}}
\toprule
    Data component & Information available & Permitted use \\
    \midrule
    Source training set
        &Covariates and labels
        &Fit preprocessing and the frozen model pool \\
    Source validation set
        &Covariates and labels 
        & Fit residual variance heads and select the source-best model \\
    Full target probe
        & Covariates only
        & Fit the partition used by the regional mechanism \\
    Diagnostic sample $\mathcal{P}_D$
        & Covariates and labels
        & Estimate $\Dcf$ by cross-fitting, then refit the static and regionwise convex combiners for C1 test evaluation \\
    Combiner sample $\mathcal{P}_C$
        & Covariates and labels
        & Fit the static floor and candidate combiners \\
    Gate sample $\mathcal{P}_G$
        & Covariates and labels
        & Validate candidates against the static floor \\
    Target test set
        & Covariates and labels
        & Final evaluation only \\
    \end{tabularx}
    \label{tab:protocol}
\end{table}

\section{Does the diagnostic predict realizable gains?}
\label{sec:c1}
We evaluate whether $\Dcf$, computed from the diagnostic probe, predicts the realized regionwise gain on the disjoint target test set. The primary analysis uses the frozen suite of 12 dataset/shift pairs: five development pairs and seven pairs from two preregistered held-out batches. Each pair is evaluated with three seeds. The dataset/shift pair is the statistical unit, so we first average over seeds within each pair and then compute all primary statistics across the 12 dataset-level means.

\subsection{Primary association}

\begin{figure}[t]
    \centering
    \includegraphics[width=0.70\linewidth]{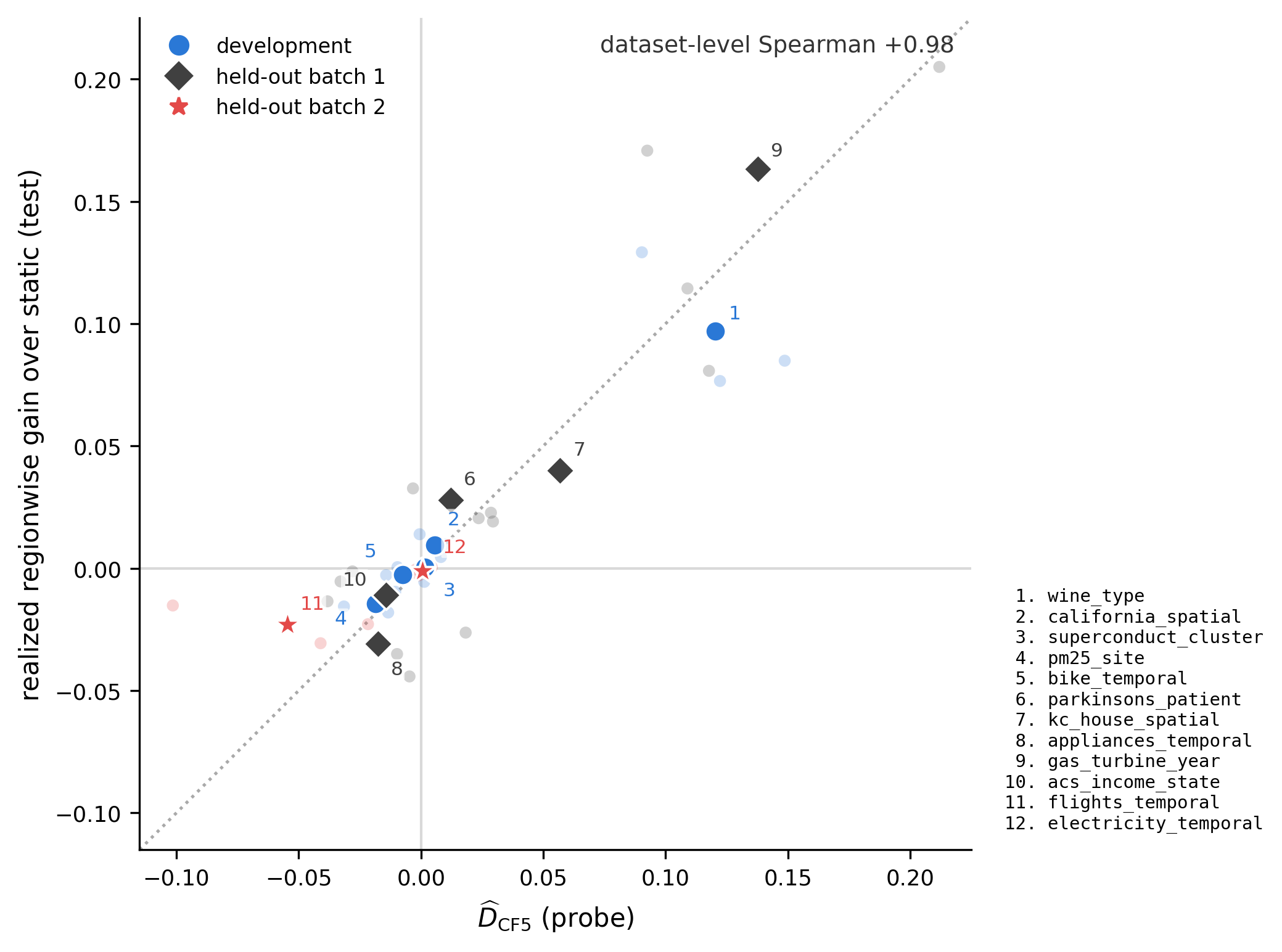}
    \caption{$\Dcf$ versus realized regionwise gain on the frozen 12-pair suite. Large markers show dataset-level means and faded markers show the three individual seeds. The dotted line is the identity reference. Dataset-level Spearman correlation is $+0.98$ (bootstrap 95\% CI $[+0.83,+1.00]$; permutation $p<10^{-4}$). Per-dataset 95\% $t_2$ intervals appear in \cref{fig:c1-with-ci}.}
    \label{fig:c1}
\end{figure}

\Cref{fig:c1} compares $\Dcf$ with realized regional gain. Their dataset-level Spearman correlation is $+0.979$, with a bootstrap 95\% confidence interval of $[+0.83,+1.00]$ and a one-sided permutation-test $p$-value of $5\times10^{-5}$. The Pearson correlation is $+0.960$, and the linear calibration of realized gain on $\Dcf$, fitted with an intercept, has slope $0.937$ and intercept $+0.0039$. Leave-one-dataset-out Spearman correlations range from $+0.973$ to $+1.000$, showing that the association is not driven by one shift.

The bootstrap and permutation procedures treat the dataset/shift pair as the resampling unit. The confidence interval is a percentile interval based on 10,000 dataset-level bootstrap resamples. The one-sided permutation test uses 20,000 permutations of the realized-gain values relative to the fixed diagnostic values, with a plus-one correction.

For the ten dataset/shift pairs whose mean realized gain lies outside the preregistered indeterminate interval of $\lvert\widehat G_{\mathrm{test}}\rvert<0.002$, the sign of $\Dcf$ matches the sign of realized gain in all ten cases. Together with the dataset-level rank correlation, this establishes a strong across-dataset ordering and the decision boundary between gains and losses. The near-unit calibration slope summarizes the overall scale of the relationship rather than exact pair-level magnitudes. For example, flights has a more negative $\Dcf$ than appliances, while appliances incurs the larger realized loss. Run-specific deployment risk is handled by the held-out validation stage in \cref{sec:selector}. 

\subsection{Comparison with simpler diagnostics}

We compare $\Dcf$ with six alternatives computed from the same diagnostic probe. Shift severity measures the increase in uniform ensemble loss relative to source validation. Prediction disagreement is the average variance of the pool predictions. The heterogeneity summary measures regional variation in uniform loss, while the competence summary measures the relative advantage of the locally best model. We also consider their product and the in-sample version of the static-versus-regional headroom contrast. 

For each diagnostic, we report its dataset-level Spearman correlation with realized gain. We additionally fit a linear calibration on 11 dataset means, predict the omitted dataset, and average the absolute leave-one-out errors. Lower leave-one-out mean absolute error indicates a more useful quantitative prediction of gain.

\begin{table}[t]
\centering
\small
\caption{Probe diagnostics on the frozen 12-pair suite. Spearman correlation is computed against realized regional gain. LOO-MAE is the leave-one-dataset-out calibration error in standardized-MSE units.}
\label{tab:diagnostic-comparison}
\begin{tabular*}{0.75\linewidth}{@{\extracolsep{\fill}}lcc@{}}
\toprule
Diagnostic & Spearman & LOO-MAE \\
\midrule
$\Dcf$                         & $+0.979$ & $0.0158$ \\
Shift severity                 & $+0.657$ & $0.0432$ \\
Heterogeneity summary          & $+0.545$ & $0.0371$ \\
Heterogeneity $\times$ competence
                               & $+0.503$ & $0.0342$ \\
Local-competence summary       & $+0.469$ & $0.0441$ \\
In-sample headroom             & $+0.357$ & $0.0259$ \\
Prediction disagreement        & $+0.021$ & $0.0517$ \\
\bottomrule
\end{tabular*}
\end{table}

No alternative reaches a correlation above $+0.66$. In particular, the in-sample version of the same contrast is less predictive than its cross-fitted counterpart. Cross-fitting therefore contributes more than a generic summary of shift magnitude, model diversity, or apparent regional flexibility.

\subsection{Held-out evidence}
The two preregistered held-out batches test whether the relationship survives beyond the datasets used to develop the diagnostic. In the first batch, our recorded expectations were wrong for two of five shifts. We expected little or negative regional gain on gas-turbine emissions, but $\Dcf$ was positive and the realized gains were the largest in the suite. We expected a weak positive gain for the ACS state shift, but $\Dcf$ was near zero or negative and the realized gain was likewise non-positive. In both cases, the probe diagnostic corrected the direction of the recorded prior.

The second held-out batch added flight-delay and electricity-load shifts without changing the frozen diagnostic or analysis protocol. Their inclusion preserved the dataset-level relationship shown in \cref{fig:c1}.

\subsection{Mechanism specificity}
The diagnostic is designed for one mechanism: reallocating convex model trust across regions. Its strong association does not imply that it predicts every form of input-dependent combination. To test this boundary, we compare $\Dcf$ with the incremental test gain of covariate-dependent stacking after accounting for static affine correction.

The correlation remains $+0.979$ for the regionwise-convex gain that $\Dcf$ targets, but falls to $+0.014$ for the incremental value of covariate-dependent stacking over static affine stacking. By contrast, measuring covariate-dependent stacking against the weaker static convex baseline produces a correlation of $+0.790$, largely because both methods can benefit from affine correction. The near-zero conditional correlation shows that $\Dcf$ isolates regional convex trust reallocation rather than acting as a generic score for all dynamic combiners.

\subsection{Sensitivity across all 16 pairs}
\label{sec:c1-sensitivity}
Four additional dataset/shift pairs came from batch 3, which was preregistered for prospective selector validation rather than C1 development. Including their C1 quantities only as a sensitivity analysis reduces the dataset-level Spearman correlation from $+0.979$ to $+0.829$, with bootstrap 95\% confidence interval $[+0.46,+1.00]$ and permutation $p=5\times10^{-5}$. The corresponding 16-pair plot appears in \cref{app:c1-sensitivity}.

The added pairs also clarify the diagnostic's operating boundary. Protein structure contains a direction error, while the online-news shift shows a more consequential magnitude error: $\Dcf$ remains near zero while unconditional regional deployment can incur a very large test loss. Thus, $\Dcf$ provides a useful ranking and directional diagnosis for the regional mechanism, while a separate validation gate is needed to control deployment-specific tail failures. This motivates the Probe-Validated Ensemble Selector in \cref{sec:selector}.

\section{When do dynamic gains exist?}
\label{sec:mechanism}

The preceding section shows that $\Dcf$ predicts realizable regionwise gain. We now isolate the conditions that create this gain and measure how shift severity and probe budget affect its realization. 

\paragraph{Controlled generator.}
Let $r_i\in\{1,\ldots,J_\star\}$ denote the generator region of observation $i$, where $J_\star=8$, and let $\bar\mu_i=K^{-1}\sum_{k=1}^K\mu_{ik}$. The generator first controls base-pool diversity and then applies region-dependent damage:
\begin{equation}
\begin{aligned}
\mu_{ik}^{(\rho)}
&=
(1-\rho)\mu_{ik}+\rho\bar{\mu}_i,\\
\widetilde{\mu}_{ik}
&=
\mu_{ik}^{(\rho)}
+
d\,\eta_{r_i}(h)
\bigl[1-\kappa S_{r_i k}(h)\bigr].
\end{aligned}
\label{eq:synthetic-shift}
\end{equation}

The function $\eta_r(h)$ controls whether the direction of the damage is shared or varies across regions, while $S_{rk}(h)$ controls which model is spared. At $h=0$, the direction is constant and the same model is spared in every region. At $h=1$, the direction alternates and the spared model rotates across regions; intermediate values use linear interpolation. The severity $d$ sets the damage magnitude, while $\kappa$ controls the degree of local sparing: $\kappa=0$ damages all models equally, whereas $\kappa=1$ applies the maximum region-specific sparing defined by $S_{rk}(h)$. Finally, $\rho$ homogenizes the original pool, so $1-\rho$ indexes generic base-pool diversity. Target labels remain unchanged. The explicit definitions of $\eta_r(h)$ and $S_{rk}(h)$, together with the complete parameter grid, appear in \cref{app:synthetic}.

The generator regions are unavailable to the combination method. The method instead learns its own $J=8$ partition from probe covariates, allowing partition estimation and approximation error to enter the realized gain.

\paragraph{Experimental design.}
We evaluate the $(h,\kappa)$ grid using 10 independently seeded model pools and three target splits per pool. Splits are averaged within each pool, making the model pool the unit of inference. Unless a control is being varied, we fix $d=0.5$, $n=512$, and $\rho=0$.

Realized gain is measured by the selector frozen before these experiments. It fits regional, smooth covariate-gated, and precision-weighted candidates on the combiner split and deploys a candidate only after validation on the held-out gate split against a static floor. The deployment procedure is introduced in \cref{sec:selector}.

\paragraph{Heterogeneity and local competence act jointly.}
The three corners missing at least one condition have negligible realized gain: $+0.0000$ at $(h,\kappa)=(0,0)$, $+0.0007$ at $(0,1)$, and $-0.0000$ at $(1,0)$, with all three 95\% confidence intervals containing zero. At $(1,1)$, realized gain reaches $+0.0910$ with 95\% CI $[+0.0844,+0.0975]$.

Within each pool, we regress gain on $h$, $\kappa$, and their interaction. The pool-level interaction is
$\beta_{h\kappa}=+0.0830$ with 95\% CI $[+0.0760,+0.0900]$ and $t_9=26.8$. The corresponding interaction is $+0.0842$ for the oracle ceiling and $+0.0837$ for $\Dcf$. The diagnostic therefore recovers the same conjunctive structure as the realized gain.

\begin{figure}[t]
\centering
\includegraphics[width=0.70\linewidth]{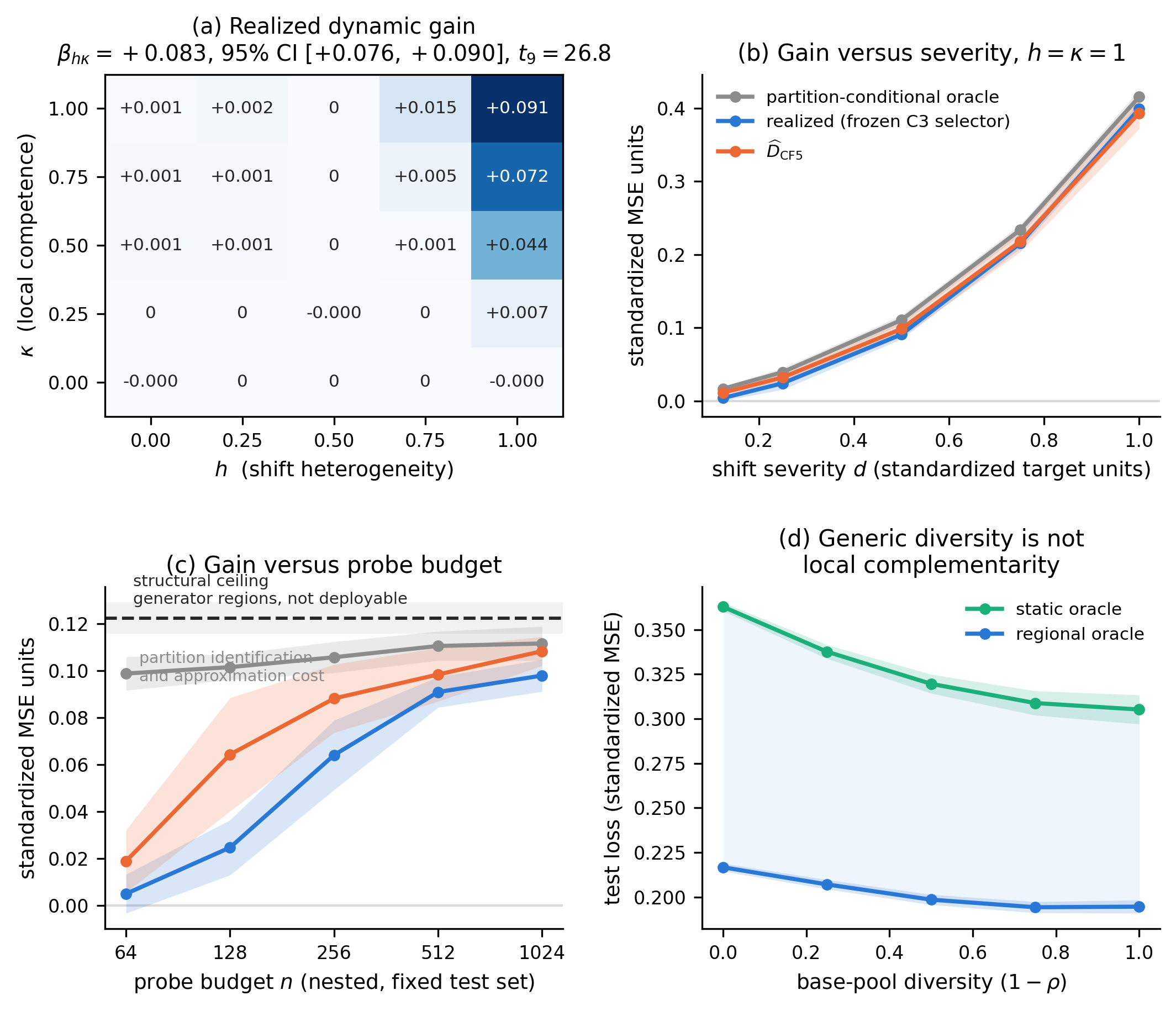}
\caption{Controlled mechanism experiments. (a) Realized gain over the $(h,\kappa)$ grid. (b) Gain as shift severity increases. (c) Gain under nested probe budgets with a fixed test set. (d) Oracle loss as generic base-pool diversity changes. Results average three target splits within each of 10 independently seeded model pools; bands show pool-level 95\% $t_9$ confidence intervals.}
\label{fig:c3}
\end{figure}

\paragraph{Severity and probe budget.}
With $h=\kappa=1$, realized gain, $\Dcf$, and the partition-conditional oracle increase together with shift severity, as shown in panel (b) of \cref{fig:c3}. Severity enlarges an available regional advantage but does not create one without heterogeneity and local competence.

The budget experiment in panel (c) of \cref{fig:c3} uses nested probes $n\in\{64,128,256,512,1024\}$ and a fixed test set. The structural ceiling, computed using the generator regions and test-label oracle weights, remains fixed at $0.1225\pm0.0067$. The partition-conditional oracle uses the probe-fitted partition and rises from $0.099$ to $0.112$ as the partition improves. Realized gain rises from $0.005$ to $0.098$. Gains are near zero at 64 labels, remain modest at 128, and become substantial at 256. The practical transition in the tested grid therefore lies between 128 and 256 labels.

\paragraph{Generic diversity is not local complementarity.}
Classical ensemble analysis relates prediction disagreement to the gain from global averaging \citep{krogh1994neural}. Our experiment separates this generic diversity from local complementarity. Increasing base-pool diversity reduces static-oracle loss by $0.058$ but regional-oracle loss by only $0.022$, as shown in panel (d) of \cref{fig:c3}. Dynamic headroom consequently narrows from $0.146$ to $0.111$. Generic diversity mainly strengthens global combination; dynamic gain requires models that remain competent in different regions.

\section{From diagnosis to deployment}
\label{sec:selector}
The diagnostic results in \cref{sec:c1} characterize realizable gain from regionwise trust allocation. The mechanism experiments in \cref{sec:mechanism} show when that gain exists. Deployment presents a broader choice: a shift may favor regional combination, global affine correction, or smooth covariate-dependent weighting. We therefore use the Probe-Validated Ensemble Selector to compare these mechanisms on held-out target data. 

\paragraph{Candidate pool and static floor.}
The selector fits a static convex blend as its deployment floor. It also fits six candidates on the combiner split $\mathcal P_C$:

\begin{enumerate}
    \item static affine stacking, $\widehat y(x)=b+\sum_{k=1}^{K}a_k\widehat f_k(x)$, with unrestricted coefficients and intercept \citep{wolpert1992stacked};
    \item CDST-RBF, which assigns affine model weights through an RBF basis of the covariates \citep{wakayama2025ensemble};
    \item a Shrunken Regional Convex Combiner;
    \item a neural mixture-of-experts gate over the frozen models \citep{jacobs1991adaptive};
    \item probe-fit inverse-variance weighting; 
    \item a linear covariate gate with softmax model weights.
\end{enumerate}

For the regional candidate, let $\widehat w_0$ be the global convex weights and $\widehat w_r$ the convex weights fitted in region $r$. The regional weights are
\begin{equation}
    \widetilde w_r
    =
    \lambda_r\widehat w_r
    +
    (1-\lambda_r)\widehat w_0,
    \qquad
    \lambda_r=\frac{n_r}{n_r+10}.
    \label{eq:selector-regional-shrinkage}
\end{equation}

A region containing fewer than five observations from $\mathcal P_C$ uses $\widehat w_0$. This shrinkage reduces the variance of regional weights estimated from small samples.

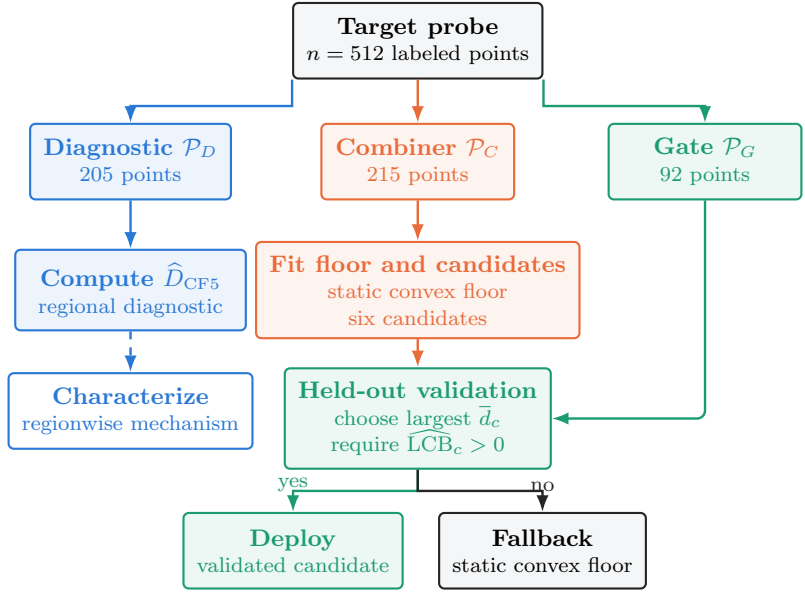
\begin{figure}[t]
    \centering
    \begin{tikzpicture}[
    x=1cm,
    y=1cm,
    box/.style={
        draw=flowink,
        rounded corners=2pt,
        line width=0.8pt,
        align=center,
        inner sep=5pt,
        minimum height=1.0cm,
        font=\small
    },
    split/.style={
        box,
        minimum width=2.55cm
    },
    operation/.style={
        box,
        minimum width=3.05cm
    },
    output/.style={
        box,
        minimum width=2.25cm
    },
    bluearrow/.style={
        -{Latex[length=2.2mm]},
        draw=flowblue,
        line width=0.9pt
    },
    orangearrow/.style={
        -{Latex[length=2.2mm]},
        draw=floworange,
        line width=0.9pt
    },
    greenarrow/.style={
        -{Latex[length=2.2mm]},
        draw=flowgreen,
        line width=0.9pt
    },
    blackarrow/.style={
        -{Latex[length=2.2mm]},
        draw=flowink,
        line width=0.9pt
    }
]

% Target probe
\node[
    box,
    fill=flowgray,
    minimum width=2.80cm
] (probe) at (0,5.4) {
    \textbf{Target probe}\\[-1pt]
    \footnotesize $n=512$ labeled points
};

% Three probe splits
\node[
    split,
    draw=flowblue,
    fill=flowblue!8,
    text=flowblue
] (diagnostic-split) at (-3.8,3.8) {
    \textbf{Diagnostic $\mathcal P_D$}\\[-1pt]
    \footnotesize 205 points
};

\node[
    split,
    draw=floworange,
    fill=floworange!8,
    text=floworange
] (combiner-split) at (0,3.8) {
    \textbf{Combiner $\mathcal P_C$}\\[-1pt]
    \footnotesize 215 points
};

\node[
    split,
    draw=flowgreen,
    fill=flowgreen!8,
    text=flowgreen
] (gate-split) at (3.8,3.8) {
    \textbf{Gate $\mathcal P_G$}\\[-1pt]
    \footnotesize 92 points
};

% Diagnostic branch
\node[
    operation,
    draw=flowblue,
    fill=flowblue!8,
    text=flowblue
] (compute-diagnostic) at (-3.8,2.1) {
    \textbf{Compute $\Dcf$}\\[-1pt]
    \footnotesize regional diagnostic
};

\node[
    operation,
    draw=flowblue,
    text=flowblue
] (characterize) at (-3.8,0.5) {
    \textbf{Characterize}\\[-1pt]
    \footnotesize regionwise mechanism
};

% Deployment branch
\node[
    operation,
    draw=floworange,
    fill=floworange!8,
    text=floworange,
    minimum width=3.35cm
] (fit-candidates) at (0,2.1) {
    \textbf{Fit floor and candidates}\\[-1pt]
    \footnotesize static convex floor\\[-1pt]
    \footnotesize six candidates
};

\node[
    operation,
    draw=flowgreen,
    fill=flowgreen!8,
    text=flowgreen,
    minimum width=3.35cm,
    minimum height=1.25cm
] (validation) at (0,0.4) {
    \textbf{Held-out validation}\\[-1pt]
    \footnotesize choose largest $\overline d_c$\\[-1pt]
    \footnotesize require
    $\widehat{\mathrm{LCB}}_c>0$
};

% Deployment outputs
\node[
    output,
    draw=flowgreen,
    fill=flowgreen!8,
    text=flowgreen
] (deploy) at (-1.65,-1.35) {
    \textbf{Deploy}\\[-1pt]
    \footnotesize validated candidate
};

\node[
    output,
    fill=flowgray
] (fallback) at (1.65,-1.35) {
    \textbf{Fallback}\\[-1pt]
    \footnotesize static convex floor
};

% Split the target probe
\draw[bluearrow]
    (probe.south west)
    -- ++(0,-0.35)
    -| (diagnostic-split.north);

\draw[orangearrow]
    (probe.south)
    -- (combiner-split.north);

\draw[greenarrow]
    (probe.south east)
    -- ++(0,-0.35)
    -| (gate-split.north);

% Diagnostic path
\draw[bluearrow]
    (diagnostic-split.south)
    -- (compute-diagnostic.north);

\draw[bluearrow,dashed]
    (compute-diagnostic.south)
    -- (characterize.north);

% Candidate-fitting path
\draw[orangearrow]
    (combiner-split.south)
    -- (fit-candidates.north);

\draw[orangearrow]
    (fit-candidates.south)
    -- (validation.north);

% Gate data enter validation separately
\draw[greenarrow, rounded corners=3pt]
    (gate-split.south)
    -- (gate-split.south |- validation.east)
    -- (validation.east);

% Final decision
\draw[greenarrow]
    (validation.south)
    -- ++(0,-0.28)
    -|
    node[
        near end,
        above,
        font=\footnotesize,
        text=flowgreen
    ] {yes}
    (deploy.north);

\draw[blackarrow]
    (validation.south)
    -- ++(0,-0.28)
    -|
    node[
        near end,
        above,
        font=\footnotesize,
        text=flowink
    ] {no}
    (fallback.north);

\end{tikzpicture}
    \caption{Diagnostic and deployment workflow. Probe labels have disjoint uses: $\mathcal P_D$ estimates $\Dcf$, $\mathcal P_C$ fits the static floor and six candidates, and $\mathcal P_G$ validates the deployment decision. The partition is learned from all probe covariates without using their labels. The diagnostic characterizes the regional mechanism but does not enter the selector rule.}
    \label{fig:selector-workflow}
\end{figure}

The three probe splits and their roles are summarized in \cref{fig:selector-workflow}.

\paragraph{Held-out validation rule.}
Let $f_0$ denote the static convex floor and $f_c$ a candidate fitted on $\mathcal P_C$. For every observation $i\in\mathcal P_G$, define the improvement of candidate $c$ over the floor as 
\begin{equation}
    d_{ic}
    =
    \ell\!\left(y_i,f_0(x_i)\right)
    -
    \ell\!\left(y_i,f_c(x_i)\right).
    \label{eq:selector-gate-improvement}
\end{equation}
Let $\overline d_c$ and $s_c$ denote the sample mean and standard deviation of these improvements on $\mathcal P_G$. The selector first chooses
\begin{equation}
    c^\star
    =
    \underset{c\in\{1,\ldots,6\}}{\arg\max}\;
    \overline d_c.
    \label{eq:selector-best-candidate}
\end{equation}
It then computes
\begin{equation}
    \widehat{\mathrm{LCB}}_{c^\star}
    =
    \overline d_{c^\star}
    -
    2.42\,
    \frac{s_{c^\star}}{\sqrt{|\mathcal P_G|}},
    \qquad
    |\mathcal P_G|=92.
    \label{eq:selector-lcb}
\end{equation}
The prespecified critical value $2.42$ approximates a one-sided 95\% Bonferroni correction for the six candidates. The selector deploys $f_{c^\star}$ when $\widehat{\mathrm{LCB}}_{c^\star}>0$. Otherwise, it deploys the static convex floor. 

\paragraph{Retrospective and prospective evaluation.}
The final selector was first evaluated retrospectively on the original 12 dataset/shift pairs. It was then frozen and evaluated on a preregistered prospective batch containing four additional pairs. Table~\ref{tab:selector-results} summarizes the two stages. We define $\Delta L=L_{\mathrm{selector}}-L_{\mathrm{floor}}$ in standardized-MSE units, so negative values favor the selector.

\begin{table}[t]
    \centering
    \small
    \caption{Evaluation of the Probe-Validated Ensemble Selector.}
    \label{tab:selector-results}
    \begin{tabular}{@{}p{0.20\linewidth}p{0.34\linewidth}p{0.36\linewidth}@{}}
    \toprule
    Stage & Decisions & Test outcome \\
    \midrule
    Retrospective,
    $12\times3$ runs
    &
    11 affine, 10 CDST-RBF, 1 regional, 14 floor
    &
    mean $\Delta L=-0.184$; worst $\Delta L=+0.000$
    \\
    \addlinespace
    Preregistered prospective,
    $4\times3$ runs
    &
    2 CDST-RBF, 10 floor
    &
    mean $\Delta L=-0.0035$; worst $\Delta L=+0.000$
    \\
    \bottomrule
    \end{tabular}
\end{table}

In the retrospective evaluation, the selector deployed a candidate in 22 of 36 runs. Static affine stacking accounted for 11 selections, CDST-RBF for 10, and the regional candidate for one. The selector used the floor in the remaining 14 runs and matched or improved it in all 36 observed runs.

The prospective batch was preregistered before the four datasets were downloaded or inspected. The selector used the floor in 10 of 12 runs and deployed CDST-RBF in two Seoul Bike runs. Both deployments reduced test loss relative to the floor by $11\%$ and $16\%$, and all 12 runs matched or improved the floor. 

\paragraph{The value of held-out validation.}
Across the 72 candidate and run comparisons in the prospective batch, 40 unconditional candidate deployments increased test loss. In one News Popularity run, unconditional CDST-RBF produced a loss $30.18$ times the static floor. The selector rejected it and retained the floor.

The Seoul Bike result illustrates the separation between diagnosis and deployment. Its $\Dcf$ was near zero, and the regional candidate was never selected. The selector instead validated CDST-RBF in two runs, identifying a smooth covariate-dependent mechanism outside the regional contrast measured by $\Dcf$.

\section{Related work}
\label{sec:related}
\paragraph{Static ensemble combination.}
Uniform averaging remains a strong baseline for independently trained predictors, including deep ensembles \citep{lakshminarayanan2017simple,ovadia2019can}. Classical ensemble analysis relates the benefit of averaging to prediction diversity \citep{krogh1994neural}. For majority-vote classification, \citet{theisen2023ensembles} characterize ensemble improvement through the disagreement-error ratio. Stacking instead estimates a global combination from held-out predictions \citep{wolpert1992stacked,breiman1996stacked}. These approaches use one set of weights across the input space. Our diagnostic measures whether relaxing this global rule region by region yields an out-of-sample gain large enough to offset finite-probe estimation; plain model disagreement alone has a correlation only $+0.02$ with that gain.

\paragraph{Input-dependent combination.}
Mixture-of-experts models learn a gate that assigns input-dependent weights to specialized predictors \citep{jacobs1991adaptive}. Dynamic selection similarly chooses an expert or subset of experts for each query, with most of the literature centered on classification \citep{cruz2018dynamic}. Dynamic regressor selection extends local competence estimation to continuous outcomes \citep{moura2019evaluating}. Covariate-dependent stacking instead models smooth weight functions over the input space \citep{wakayama2025ensemble}. These methods provide different mechanisms for realizing local specialization. HyRe instead uses a small labeled target sample to globally reweight ensemble heads  through a generalized Bayesian update, including in regression under covariate shift \citep{lee2026hypothesis}. Its weights are shared across target inputs. $\Dcf$ addresses the preceding decision: whether regionwise trust reallocation is supported by the available target probe. The selector in \cref{sec:selector} then compares that regional mechanism with affine and smooth covariate-dependent alternatives.

\paragraph{Uncertainty-based weighting.}
Regression models can learn input-dependent predictive variances together with their conditional means \citep{nix1994estimating,kendall2017uncertainties}. These estimates motivate precision weighting, $w_k(x)\propto1/\widehat{\sigma}_k^2(x)$. Its effectiveness under shift depends on whether the estimated variances continue to rank target errors. In our experiments, source-fitted precision weighting increased California Housing test MSE by approximately $50\%$ relative to uniform averaging. We therefore evaluate source-fitted and probe-fitted variants, while the selector admits the probe-fitted variant only after held-out validation.

\paragraph{Model evaluation under distribution shift.}
Distribution shift can degrade both predictive accuracy and uncertainty estimates \citep{ovadia2019can}, motivating benchmarks that preserve the structure of naturally occurring shifts \citep{koh2021wilds}. Under covariate shift, importance-weighted cross-validation estimates target risk by reweighting source observations with a target-to-source density ratio \citep{sugiyama2007covariate}. Our setting instead provides a small labeled sample from the target domain. This permits direct target-domain cross-fitting without assuming that the shift is purely covariate shift or estimating a density ratio. Test-time adaptation instead updates a deployed model using unlabeled target batches. Representative methods such as TENT use target-batch normalization statistics and update affine normalization parameters by minimizing predictive entropy \citep{wang2021tent}. Our protocol keeps the source-trained regression models frozen and exposes only their predictions to the combination layer, so methods requiring access to model internals fall outside the evaluation scope. OpenRegShift instead evaluates regression model combination across spatial, temporal, domain, and feature-cluster shifts.

\paragraph{Cross-fitting and held-out validation.}
Cross-validation estimates out-of-sample risk by separating fitting from evaluation \citep{arlot2010survey}. Cross-fitting extends this separation across folds and is central to double and debiased machine learning \citep{chernozhukov2018double}. We apply the same principle to a paired predictive-risk contrast: both combination classes are fitted away from each observation used to compute $\Dcf$. The diagnostic split $\mathcal P_D$ is separate from the gate split $\mathcal P_G$, which validates deployment against the static convex floor.

\section{Limitations and conclusion}
\label{sec:conclusion}
\paragraph{Scope of the diagnostic.}
$\Dcf$ estimates one specific contrast: the realizable gain of regionwise convex combination over static convex blending under a probe-fitted partition. It does not measure every form of input-dependent adaptation. Smooth covariate-dependent weighting can provide gains when $\Dcf$ is near zero, as the Seoul Bike result in \cref{sec:selector} demonstrates. Extending the diagnostic to several structured mechanism classes, while preserving out-of-sample estimation, is a natural next step.

The current implementation fixes $J=8$, uses hard regions, and evaluates standardized squared error. These choices make the estimand transparent and align it with the deployed regional candidate, but they may miss
structure at another resolution or across overlapping regions. Cross-fitted selection over multiple partitions, soft partitions, and alternative regression losses could broaden the diagnostic without changing its underlying risk-comparison principle.

\paragraph{Probe size and deployment validation.}
The method requires labeled target observations. The controlled experiments in \cref{sec:mechanism} show that the finite-probe cost is substantial: gains are small at 64 and 128 labels, with the transition occurring between 128 and 256 labels in the tested grid. Applications with smaller probes may require stronger structural assumptions, information sharing across regions, or sequential acquisition of target labels. 

The selector's lower confidence bound is an asymptotic validation rule based on the held-out gate split. Its prospective 12-run evaluation matched or improved the static floor in every run, but this empirical result is not a finite-sample guarantee. Truncating the loss differences would permit empirical-Bernstein or betting-based bounds for bounded means, providing finite-sample control for the truncated risk contrast \citep{waudbysmith2024estimating}.

\paragraph{Empirical scope.}
The frozen primary analysis contains 12 dataset/shift pairs, with four additional pairs used for sensitivity and prospective selector validation. Each real-data experiment uses three seeds; the synthetic study uses 10 independently generated model pools with three splits per pool. The two held-out C1 batches and the prospective selector batch were within-project preregistrations: the datasets were selected by the same research team, but the hypotheses, protocols, and acceptance criteria were frozen before evaluation. The suite covers spatial, temporal, domain, and feature-cluster shifts, primarily in tabular regression with frozen source-trained model pools. Evaluation on high-dimensional inputs, larger pools, additional model families, and shifts collected from deployed systems would test how broadly the characterization extends.

\paragraph{Conclusion.}
Dynamic ensembling pays when model failures vary across the input space and useful local competence remains. $\Dcf$ makes this regional opportunity measurable from a labeled target probe. Across the frozen suite, it predicts the sign and ordering of realizable regionwise gains, the controlled experiments identify the interaction that produces those gains, and the Probe-Validated Ensemble Selector chooses among complementary mechanisms while retaining a static convex fallback. OpenRegShift packages these components into a reproducible protocol for studying regression ensembles under distribution shift. 

\bibliographystyle{plainnat}
\bibliography{refs}

\begin{thebibliography}{31}
\providecommand{\natexlab}[1]{#1}
\providecommand{\url}[1]{\texttt{#1}}
\expandafter\ifx\csname urlstyle\endcsname\relax
  \providecommand{\doi}[1]{doi: #1}\else
  \providecommand{\doi}{doi: \begingroup \urlstyle{rm}\Url}\fi

\bibitem[Arlot and Celisse(2010)]{arlot2010survey}
Sylvain Arlot and Alain Celisse.
\newblock A survey of cross-validation procedures for model selection.
\newblock \emph{Statistics Surveys}, 4:\penalty0 40--79, 2010.
\newblock \doi{10.1214/09-SS054}.
\newblock URL \url{https://doi.org/10.1214/09-SS054}.

\bibitem[Breiman(1996)]{breiman1996stacked}
Leo Breiman.
\newblock Stacked regressions.
\newblock \emph{Machine Learning}, 24:\penalty0 49--64, 1996.
\newblock \doi{10.1007/BF00117832}.
\newblock URL \url{https://doi.org/10.1007/BF00117832}.

\bibitem[Chernozhukov et~al.(2018)Chernozhukov, Chetverikov, Demirer, Duflo, Hansen, Newey, and Robins]{chernozhukov2018double}
Victor Chernozhukov, Denis Chetverikov, Mert Demirer, Esther Duflo, Christian Hansen, Whitney Newey, and James Robins.
\newblock Double/debiased machine learning for treatment and structural parameters.
\newblock \emph{The Econometrics Journal}, 21\penalty0 (1):\penalty0 C1--C68, 2018.
\newblock \doi{10.1111/ectj.12097}.
\newblock URL \url{https://doi.org/10.1111/ectj.12097}.

\bibitem[Cortez et~al.(2009)Cortez, Cerdeira, Almeida, Matos, and Reis]{cortez2009modeling}
Paulo Cortez, Ant{\'o}nio Cerdeira, Fernando Almeida, Telmo Matos, and Jos{\'e} Reis.
\newblock Modeling wine preferences by data mining from physicochemical properties.
\newblock \emph{Decision Support Systems}, 47\penalty0 (4):\penalty0 547--553, 2009.
\newblock \doi{10.1016/j.dss.2009.05.016}.
\newblock URL \url{https://doi.org/10.1016/j.dss.2009.05.016}.

\bibitem[Cruz et~al.(2018)Cruz, Sabourin, and Cavalcanti]{cruz2018dynamic}
Rafael M.~O. Cruz, Robert Sabourin, and George D.~C. Cavalcanti.
\newblock Dynamic classifier selection: Recent advances and perspectives.
\newblock \emph{Information Fusion}, 41:\penalty0 195--216, 2018.
\newblock \doi{10.1016/j.inffus.2017.09.010}.
\newblock URL \url{https://doi.org/10.1016/j.inffus.2017.09.010}.

\bibitem[Ding et~al.(2021)Ding, Hardt, Miller, and Schmidt]{ding2021retiring}
Frances Ding, Moritz Hardt, John~P. Miller, and Ludwig Schmidt.
\newblock Retiring adult: New datasets for fair machine learning.
\newblock In \emph{Advances in Neural Information Processing Systems}, volume~34, pages 6478--6490, 2021.

\bibitem[Fanaee-T and Gama(2014)]{fanaee2014event}
Hadi Fanaee-T and Jo{\~a}o Gama.
\newblock Event labeling combining ensemble detectors and background knowledge.
\newblock \emph{Progress in Artificial Intelligence}, 2\penalty0 (2--3):\penalty0 113--127, 2014.
\newblock \doi{10.1007/s13748-013-0040-3}.
\newblock URL \url{https://doi.org/10.1007/s13748-013-0040-3}.

\bibitem[Fernandes et~al.(2015)Fernandes, Vinagre, and Cortez]{fernandes2015proactive}
Kelwin Fernandes, Pedro Vinagre, and Paulo Cortez.
\newblock A proactive intelligent decision support system for predicting the popularity of online news.
\newblock In \emph{Progress in Artificial Intelligence: 17th Portuguese Conference on Artificial Intelligence ({EPIA} 2015)}, Lecture Notes in Computer Science, pages 535--546. Springer, 2015.
\newblock \doi{10.1007/978-3-319-23485-4_53}.
\newblock URL \url{https://doi.org/10.1007/978-3-319-23485-4_53}.

\bibitem[Hamidieh(2018)]{hamidieh2018data}
Kam Hamidieh.
\newblock A data-driven statistical model for predicting the critical temperature of a superconductor.
\newblock \emph{Computational Materials Science}, 154:\penalty0 346--354, 2018.
\newblock \doi{10.1016/j.commatsci.2018.07.052}.
\newblock URL \url{https://doi.org/10.1016/j.commatsci.2018.07.052}.

\bibitem[Jacobs et~al.(1991)Jacobs, Jordan, Nowlan, and Hinton]{jacobs1991adaptive}
Robert~A. Jacobs, Michael~I. Jordan, Steven~J. Nowlan, and Geoffrey~E. Hinton.
\newblock Adaptive mixtures of local experts.
\newblock \emph{Neural Computation}, 1991.
\newblock \doi{10.1162/neco.1991.3.1.79}.
\newblock URL \url{https://doi.org/10.1162/neco.1991.3.1.79}.

\bibitem[Kelly et~al.(2023)Kelly, Longjohn, and Nottingham]{kelly2023uci}
Markelle Kelly, Rachel Longjohn, and Kolby Nottingham.
\newblock The {UCI} machine learning repository, 2023.
\newblock URL \url{https://archive.ics.uci.edu}.

\bibitem[Kendall and Gal(2017)]{kendall2017uncertainties}
Alex Kendall and Yarin Gal.
\newblock What uncertainties do we need in bayesian deep learning for computer vision?
\newblock In \emph{Advances in Neural Information Processing Systems}, volume~30, 2017.
\newblock URL \url{https://proceedings.neurips.cc/paper/2017/hash/2650d6089a6d640c5e85b2b88265dc2b-Abstract.html}.

\bibitem[Koh et~al.(2021)Koh, Sagawa, Marklund, Xie, Zhang, Balsubramani, Hu, Yasunaga, Phillips, Gao, Lee, David, Stavness, Guo, Earnshaw, Haque, Beery, Leskovec, Kundaje, Pierson, Levine, Finn, and Liang]{koh2021wilds}
Pang~Wei Koh, Shiori Sagawa, Henrik Marklund, Sang~Michael Xie, Marvin Zhang, Akshay Balsubramani, Weihua Hu, Michihiro Yasunaga, Richard~Lanas Phillips, Irena Gao, Tony Lee, Etienne David, Ian Stavness, Wei Guo, Berton Earnshaw, Imran Haque, Sara~M. Beery, Jure Leskovec, Anshul Kundaje, Emma Pierson, Sergey Levine, Chelsea Finn, and Percy Liang.
\newblock {WILDS}: A benchmark of in-the-wild distribution shifts.
\newblock In \emph{Proceedings of the 38th International Conference on Machine Learning}, volume 139 of \emph{Proceedings of Machine Learning Research}, pages 5637--5664. PMLR, 2021.
\newblock URL \url{https://proceedings.mlr.press/v139/koh21a.html}.

\bibitem[Krogh and Vedelsby(1994)]{krogh1994neural}
Anders Krogh and Jesper Vedelsby.
\newblock Neural network ensembles, cross validation, and active learning.
\newblock In \emph{Advances in Neural Information Processing Systems}, volume~7, pages 231--238. MIT Press, 1994.
\newblock URL \url{https://proceedings.neurips.cc/paper/1994/hash/b8c37e33defde51cf91e1e03e51657da-Abstract.html}.

\bibitem[Lakshminarayanan et~al.(2017)Lakshminarayanan, Pritzel, and Blundell]{lakshminarayanan2017simple}
Balaji Lakshminarayanan, Alexander Pritzel, and Charles Blundell.
\newblock Simple and scalable predictive uncertainty estimation using deep ensembles.
\newblock In \emph{Advances in Neural Information Processing Systems}. Curran Associates, Inc., 2017.

\bibitem[Lee et~al.(2026)Lee, Williams, Marklund, Sharma, Mitchell, Singh, and Finn]{lee2026hypothesis}
Yoonho Lee, Jonathan Williams, Henrik Marklund, Archit Sharma, Eric Mitchell, Anikait Singh, and Chelsea Finn.
\newblock Inference-time alignment via hypothesis reweighting.
\newblock \emph{Transactions on Machine Learning Research}, 2026.
\newblock URL \url{https://openreview.net/forum?id=Q9p8LSEpiJ}.

\bibitem[Moura et~al.(2019)Moura, Cavalcanti, and Oliveira]{moura2019evaluating}
Thiago J.~M. Moura, George D.~C. Cavalcanti, and Luiz~S. Oliveira.
\newblock Evaluating competence measures for dynamic regressor selection.
\newblock In \emph{2019 International Joint Conference on Neural Networks}. IEEE, 2019.
\newblock \doi{10.1109/IJCNN.2019.8851835}.
\newblock URL \url{https://doi.org/10.1109/IJCNN.2019.8851835}.

\bibitem[Nix and Weigend(1994)]{nix1994estimating}
David~A. Nix and Andreas~S. Weigend.
\newblock Estimating the mean and variance of the target probability distribution.
\newblock In \emph{Proceedings of the 1994 IEEE International Conference on Neural Networks}, volume~1, pages 55--60. IEEE, 1994.
\newblock \doi{10.1109/ICNN.1994.374138}.
\newblock URL \url{https://doi.org/10.1109/ICNN.1994.374138}.

\bibitem[Ovadia et~al.(2019)Ovadia, Fertig, Ren, Nado, Sculley, Nowozin, Dillon, Lakshminarayanan, and Snoek]{ovadia2019can}
Yaniv Ovadia, Emily Fertig, Jie Ren, Zachary Nado, D.~Sculley, Sebastian Nowozin, Joshua~V. Dillon, Balaji Lakshminarayanan, and Jasper Snoek.
\newblock Can you trust your model's uncertainty? evaluating predictive uncertainty under dataset shift.
\newblock In \emph{Advances in Neural Information Processing Systems}. Curran Associates, Inc., 2019.

\bibitem[Pace and Barry(1997)]{pace1997sparse}
R.~Kelley Pace and Ronald Barry.
\newblock Sparse spatial autoregressions.
\newblock \emph{Statistics \& Probability Letters}, 33\penalty0 (3):\penalty0 291--297, 1997.
\newblock \doi{10.1016/S0167-7152(96)00140-X}.

\bibitem[Pedregosa et~al.(2011)Pedregosa, Varoquaux, Gramfort, Michel, Thirion, Grisel, Blondel, Prettenhofer, Weiss, Dubourg, VanderPlas, Passos, Cournapeau, Brucher, Perrot, and Duchesnay]{pedregosa2011scikit}
Fabian Pedregosa, Ga{\"e}l Varoquaux, Alexandre Gramfort, Vincent Michel, Bertrand Thirion, Olivier Grisel, Mathieu Blondel, Peter Prettenhofer, Ron Weiss, Vincent Dubourg, Jake VanderPlas, Alexandre Passos, David Cournapeau, Matthieu Brucher, Matthieu Perrot, and {\'E}douard Duchesnay.
\newblock Scikit-learn: Machine learning in python.
\newblock \emph{Journal of Machine Learning Research}, 12:\penalty0 2825--2830, 2011.

\bibitem[Sugiyama et~al.(2007)Sugiyama, Krauledat, and M{\"u}ller]{sugiyama2007covariate}
Masashi Sugiyama, Matthias Krauledat, and Klaus-Robert M{\"u}ller.
\newblock Covariate shift adaptation by importance weighted cross validation.
\newblock \emph{Journal of Machine Learning Research}, 8\penalty0 (35):\penalty0 985--1005, 2007.
\newblock URL \url{https://www.jmlr.org/papers/v8/sugiyama07a.html}.

\bibitem[Theisen et~al.(2023)Theisen, Kim, Yang, Hodgkinson, and Mahoney]{theisen2023ensembles}
Ryan Theisen, Hyunsuk Kim, Yaoqing Yang, Liam Hodgkinson, and Michael~W. Mahoney.
\newblock When are ensembles really effective?
\newblock In \emph{Advances in Neural Information Processing Systems}, volume~36. Curran Associates, Inc., 2023.
\newblock \doi{10.52202/075280-0659}.
\newblock URL \url{https://proceedings.neurips.cc/paper_files/paper/2023/hash/30b6fa308e62ed52180c31ae3ba6bb0a-Abstract-Conference.html}.

\bibitem[Tsanas et~al.(2010)Tsanas, Little, McSharry, and Ramig]{tsanas2010accurate}
Athanasios Tsanas, Max~A. Little, Patrick~E. McSharry, and Lorraine~O. Ramig.
\newblock Accurate telemonitoring of {Parkinson}'s disease progression by noninvasive speech tests.
\newblock \emph{IEEE Transactions on Biomedical Engineering}, 57\penalty0 (4):\penalty0 884--893, 2010.
\newblock \doi{10.1109/TBME.2009.2036000}.
\newblock URL \url{https://doi.org/10.1109/TBME.2009.2036000}.

\bibitem[Vanschoren et~al.(2013)Vanschoren, van Rijn, Bischl, and Torgo]{vanschoren2013openml}
Joaquin Vanschoren, Jan~N. van Rijn, Bernd Bischl, and Luis Torgo.
\newblock {OpenML}: Networked science in machine learning.
\newblock \emph{ACM SIGKDD Explorations Newsletter}, 15\penalty0 (2):\penalty0 49--60, 2013.
\newblock \doi{10.1145/2641190.2641198}.

\bibitem[Wakayama and Sugasawa(2025)]{wakayama2025ensemble}
Tomoya Wakayama and Shonosuke Sugasawa.
\newblock Ensemble prediction via covariate-dependent stacking.
\newblock \emph{Statistics and Computing}, 35\penalty0 (6):\penalty0 212, 2025.
\newblock \doi{10.1007/s11222-025-10739-y}.
\newblock URL \url{https://doi.org/10.1007/s11222-025-10739-y}.

\bibitem[Wang et~al.(2021)Wang, Shelhamer, Liu, Olshausen, and Darrell]{wang2021tent}
Dequan Wang, Evan Shelhamer, Shaoteng Liu, Bruno Olshausen, and Trevor Darrell.
\newblock {Tent}: Fully test-time adaptation by entropy minimization.
\newblock In \emph{International Conference on Learning Representations}, 2021.
\newblock URL \url{https://openreview.net/forum?id=uXl3bZLkr3c}.

\bibitem[Waudby-Smith and Ramdas(2024)]{waudbysmith2024estimating}
Ian Waudby-Smith and Aaditya Ramdas.
\newblock Estimating means of bounded random variables by betting.
\newblock \emph{Journal of the Royal Statistical Society Series B: Statistical Methodology}, 86\penalty0 (1):\penalty0 1--27, 2024.
\newblock \doi{10.1093/jrsssb/qkad009}.
\newblock URL \url{https://doi.org/10.1093/jrsssb/qkad009}.

\bibitem[Wickham(2022)]{wickham2022nycflights13}
Hadley Wickham.
\newblock \emph{nycflights13: Flights that Departed NYC in 2013}, 2022.
\newblock URL \url{https://github.com/hadley/nycflights13}.
\newblock R package version 1.0.2.

\bibitem[Wolpert(1992)]{wolpert1992stacked}
David~H. Wolpert.
\newblock Stacked generalization.
\newblock \emph{Neural Networks}, 1992.
\newblock \doi{10.1016/S0893-6080(05)80023-1}.
\newblock URL \url{https://doi.org/10.1016/S0893-6080(05)80023-1}.

\bibitem[Zhang et~al.(2017)Zhang, Guo, Dong, He, Xu, and Chen]{zhang2017cautionary}
Shuyi Zhang, Bin Guo, Anlan Dong, Jing He, Ziping Xu, and Song~Xi Chen.
\newblock Cautionary tales on air-quality improvement in {Beijing}.
\newblock \emph{Proceedings of the Royal Society A: Mathematical, Physical and Engineering Sciences}, 473\penalty0 (2205):\penalty0 20170457, 2017.
\newblock \doi{10.1098/rspa.2017.0457}.
\newblock URL \url{https://doi.org/10.1098/rspa.2017.0457}.

\end{thebibliography}

\clearpage
\appendix

\counterwithin{figure}{section}
\counterwithin{table}{section}
\counterwithin{equation}{section}

\section{Additional C1 results}
\label{app:c1_uncertainty}

\subsection{Variation across seeds}
\label{app:c1-seeds}
\begin{figure}[h!]
\centering
\includegraphics[width=0.70\linewidth]{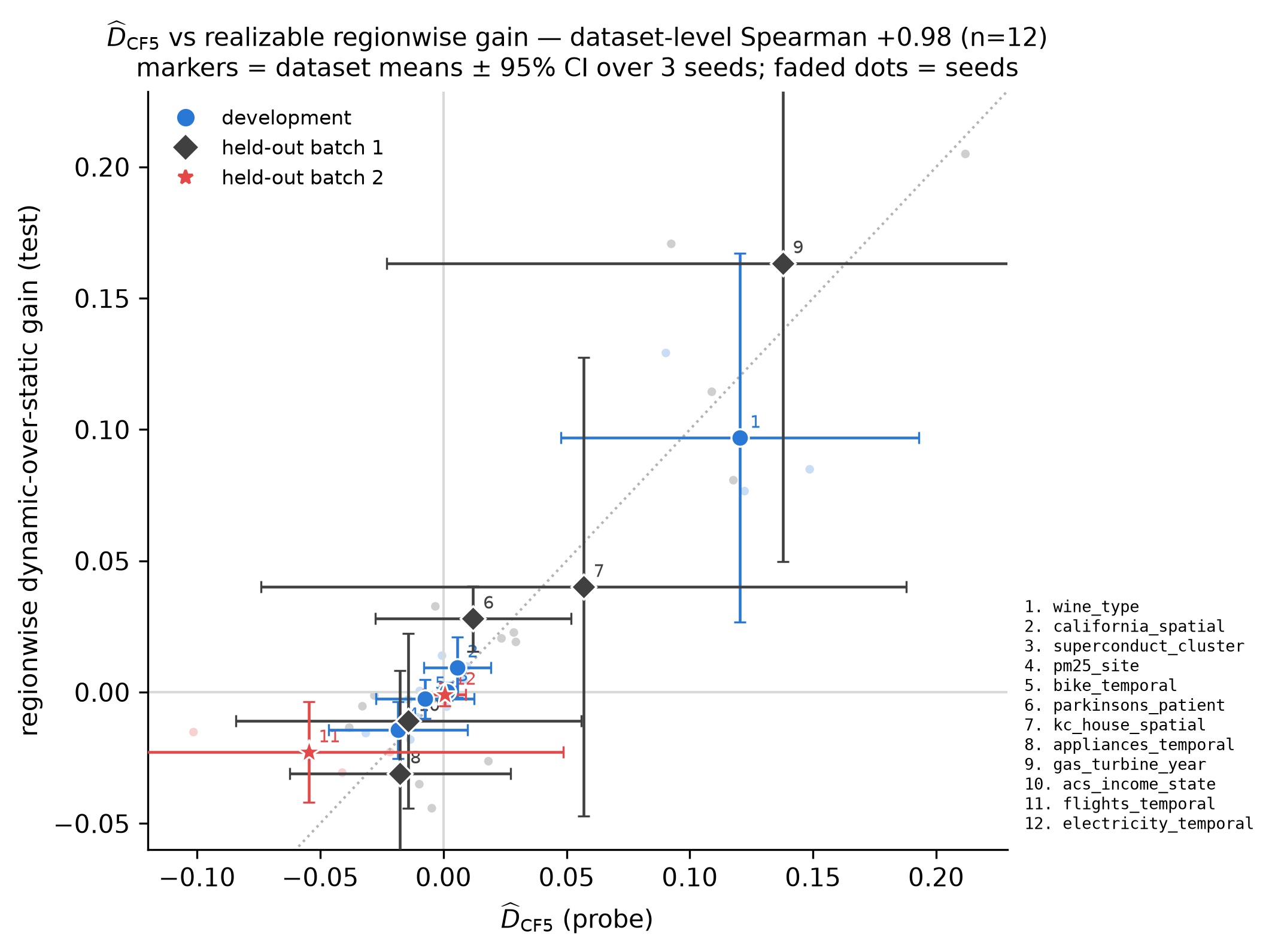}
\caption{The C1 association with per-dataset 95\% $t_2$ confidence intervals over three seeds. Large markers show dataset means and faded markers show the individual seeds. Statistical inference in the primary analysis treats the 12 dataset/shift pairs, rather than the 36 seed runs, as the independent units.}
\label{fig:c1-with-ci}
\end{figure}

The primary C1 analysis averages the three seeds within each dataset/shift pair and treats the 12 pairs as the independent units. \Cref{fig:c1-with-ci} reports per-dataset 95\% $t_2$ confidence intervals. With only three seeds, $t_{2,0.975}=4.303$, so these intervals are necessarily wide. The individual seed results are shown directly as faded points.

\subsection{Sensitivity across all 16 pairs}
\label{app:c1-sensitivity}
Batch 3 was reserved for prospective selector validation (\cref{app:preregistration}); its four pairs enter C1 only through this sensitivity analysis. Across all 16 pairs, the dataset-level Spearman correlation is $+0.829$ with bootstrap 95\% confidence interval $[+0.46,+1.00]$ and permutation $p=5\times10^{-5}$, compared with $+0.979$ on the frozen 12-pair suite. Batch 3 alone has correlation $+0.40$; with only four pairs, we report this value without further interpretation.

\begin{figure}[t]
\centering
\includegraphics[width=0.70\linewidth]{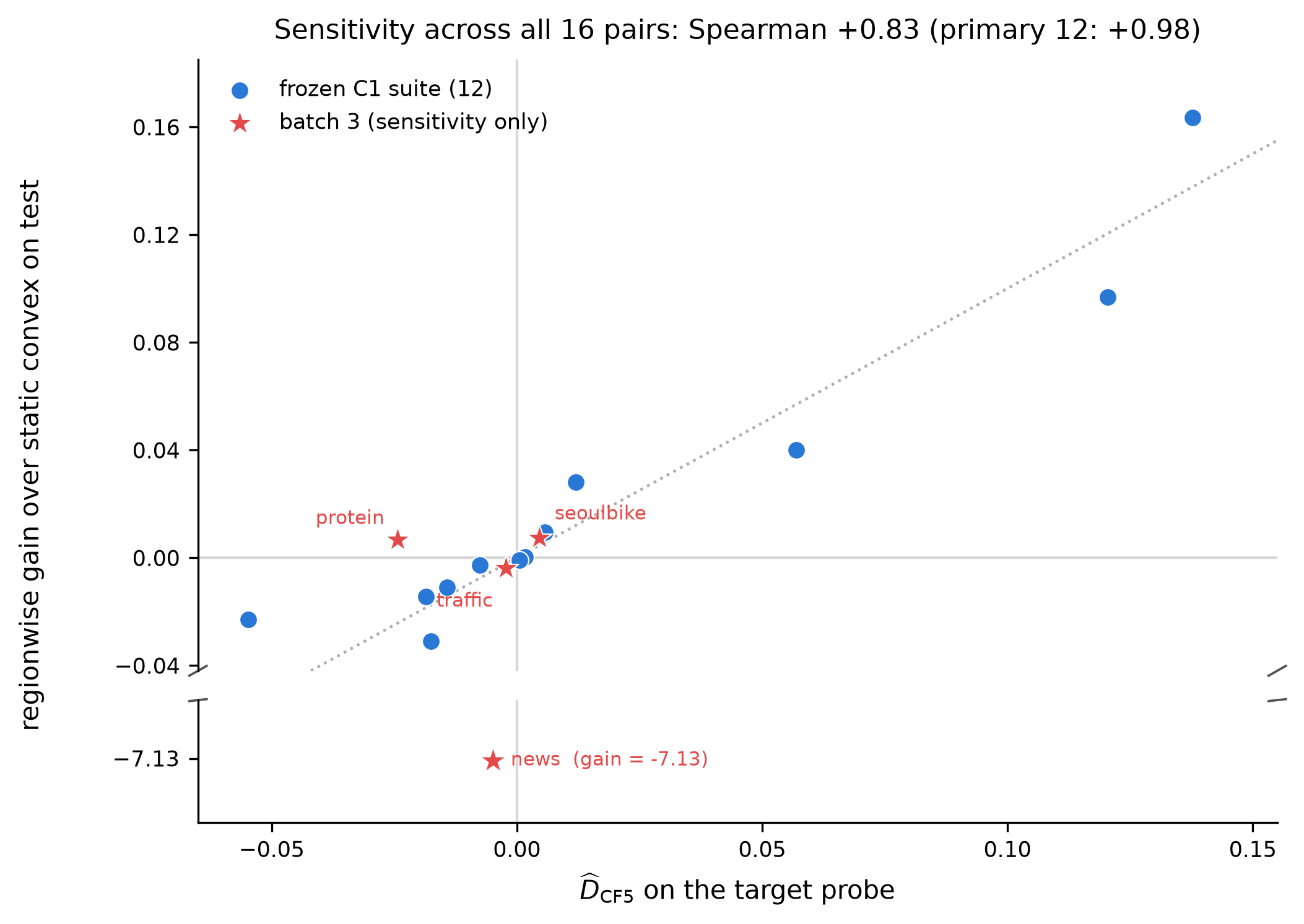}
\caption{Sensitivity analysis across all 16 dataset/shift pairs. Markers show three-seed dataset means. Blue circles denote the frozen 12-pair C1 suite and red stars denote the four pairs from the prospective selector batch. The broken vertical axis preserves the extreme News result while resolving the remaining 15 pairs.}
\label{fig:c1-sensitivity16}
\end{figure}

The News pair illustrates a limitation in magnitude prediction: $\Dcf$ is slightly negative, correctly indicating no regional opportunity, but does not anticipate the $-7.13$ standardized-MSE loss from unconditional regional deployment. Protein produces a directional miss. These cases motivate the independent deployment validation in \cref{sec:selector}.

\FloatBarrier
\subsection{Sensitivity to partition resolution}
\label{app:jsweep}
The primary experiments fix the number of regions at $J=8$. This choice was made using only the five development pairs, before evaluating either held-out C1 batch. We examined $J\in\{2,4,8,16,32\}$ while keeping the probe budget, cross-fitting procedure, and regional fallback rule unchanged. For every value of $J$, both $\Dcf$ and realized test gain refer to the corresponding partition resolution.

\Cref{tab:j-sweep} reports two descriptive correlations. The pooled correlation treats the 15 dataset-by-seed runs as observations. The dataset-level correlation first averages the three seeds, leaving five development datasets, and is therefore included only as a coarse summary. Here, $J_{\mathrm{eff}}$ is the number of regions containing at least five diagnostic observations.

\begin{table}[t]
    \centering
    \caption{Development-only sensitivity to partition resolution. Spearman correlations compare $\Dcf$ with the realized test gain of the regionwise convex combiner constructed at the same $J$.}
    \label{tab:j-sweep}
    \small
    \setlength{\tabcolsep}{8pt}
    \renewcommand{\arraystretch}{1.2}
    \begin{tabular}{@{}rrrr@{}}
    \toprule
    $J$ &
    mean $J_{\mathrm{eff}}$ &
    pooled Spearman &
    dataset-level Spearman \\
    \midrule
    2  & 1.8  & $+0.377$ & $+0.600$ \\
    4  & 3.3  & $+0.575$ & $+0.700$ \\
    8  & 7.1  & $+0.807$ & $+1.000$ \\
    16 & 13.5 & $+0.886$ & $+1.000$ \\
    32 & 19.7 & $+0.761$ & $+0.900$ \\
    \bottomrule
    \end{tabular}
\end{table}

The association remains positive throughout the sweep. Small values of $J$ provide limited partition resolution, while large values divide the 205-point diagnostic sample among many regions. Both $J=8$ and $J=16$ order the five dataset means, but $J=8$ retains twice as many diagnostic observations per requested region and requires fewer regional weight vectors. We froze $J=8$ as the tradeoff used in all subsequent held-out experiments. At $J=32$, the decline in pooled correlation is consistent with the finite-probe estimation cost that $\Dcf$ is designed to capture. 

\FloatBarrier
\section{Dataset and protocol details}
\label{app:data-protocol}

\subsection{Datasets and shift construction}
\label{app:datasets}
OpenRegShift contains 16 public dataset/shift pairs. Twelve datasets come from the UCI Machine Learning Repository \citep{kelly2023uci}. Where an introducing publication exists, we cite it: Bike \citep{fanaee2014event}, Superconduct \citep{hamidieh2018data}, PM2.5 \citep{zhang2017cautionary}, Wine \citep{cortez2009modeling}, Parkinsons \citep{tsanas2010accurate}, and News \citep{fernandes2015proactive}; the other six UCI datasets are cited through the repository entry. The remaining data come from California Housing \citep{pace1997sparse}, OpenML \citep{vanschoren2013openml}, Folktables \citep{ding2021retiring}, and \texttt{nycflights13} \citep{wickham2022nycflights13}. \Cref{tab:dataset-sources} lists their outcomes and experimental roles.

The frozen C1 analysis contains five development pairs, five pairs from held-out batch 1, and two pairs from held-out batch 2. Batch 3 contains four pairs preregistered for prospective selector validation. These four pairs enter the 16-pair C1 sensitivity analysis but not the frozen 12-pair C1 claim.

\begin{table}[t]
    \centering
    \caption{Dataset sources, prediction outcomes, and experimental roles. Dev denotes development; H1 and H2 denote the two held-out C1 batches; P3 denotes the prospective selector batch. Outcome transformations precede standardization by source-training statistics.}
    \footnotesize
    \begin{tabularx}{\linewidth}{@{}
    >{\raggedright\arraybackslash}p{0.19\linewidth}
    >{\raggedright\arraybackslash}p{0.19\linewidth}
    >{\raggedright\arraybackslash}Xl@{}}
    \toprule
    Dataset & Source & Outcome & Role \\
    \midrule
    California
        & sklearn/StatLib
        & median house value
        & Dev \\
    Bike
        & UCI 275
        & $\log(1+\text{rental count})$
        & Dev \\
    Superconduct
        & UCI 464
        & critical temperature
        & Dev \\
    PM2.5
        & UCI 501
        & $\log(1+\mathrm{PM}_{2.5})$
        & Dev \\
    Wine
        & UCI 186
        & wine quality
        & Dev \\
    Parkinsons
        & UCI 189
        & total UPDRS
        & H1 \\
    KC House
        & OpenML 42092
        & $\log(1+\text{price})$
        & H1 \\
    Appliances
        & UCI 374
        & $\log(1+\text{energy use})$
        & H1 \\
    Gas Turbine
        & UCI 551
        & $\mathrm{NO_x}$ emission
        & H1 \\
    ACS Income
        & Folktables/ACS
        & $\log(1+\mathrm{PINCP})$
        & H1 \\
    Flights
        & \texttt{nycflights13}
        & signed $\log(1+|\text{arrival delay}|)$
        & H2 \\
    Electricity
        & UCI 321
        & log aggregate load
        & H2 \\
    Protein
        & UCI 265
        & RMSD
        & P3 \\
    News
        & UCI 332
        & $\log(1+\text{shares})$
        & P3 \\
    Traffic
        & UCI 492
        & traffic volume
        & P3 \\
    Seoul Bike
        & UCI 560
        & $\log(1+\text{rental count})$
        & P3 \\
    \bottomrule
    \end{tabularx}
    \label{tab:dataset-sources}
\end{table}
The suite includes spatial, temporal, domain, and feature-cluster shifts. Every temporal split uses earlier observations as source data and later observations as target data. No future observations enter source-model training. Spatial splits separate locations, while domain splits separate populations or data types. For each feature-cluster shift, $k$-means with $k=5$, 10 initializations, and random seed 0 is applied to standardized covariates. Among clusters containing at least 2000 observations, the target is the cluster whose centroid has the largest sum of Euclidean distances to the other four centroids. Target labels do not enter this construction. 
\Cref{tab:dataset-shifts} reports the shift definitions and sample sizes used by the experiments. Here, $n_S$ includes source training and source validation observations, and $n_T$ denotes the target sample before allocating the 512-point probe. PM2.5, ACS Income, and Flights use pre-specified sample caps for computational control.

\begin{table}[t]
\centering
\caption{Shift construction and experimental sample sizes. The
512-point target probe is drawn from the $n_T$ observations; the
remaining target observations form the test set.}
\label{tab:dataset-shifts}
\footnotesize
\setlength{\tabcolsep}{3pt}
\renewcommand{\arraystretch}{1.08}
\begin{tabularx}{\linewidth}{@{}
    >{\raggedright\arraybackslash}p{0.17\linewidth}
    >{\raggedright\arraybackslash}X
    r
    r
    >{\raggedright\arraybackslash}p{0.16\linewidth}@{}}
    \toprule
    Dataset & Source $\rightarrow$ target & $n_S$ & $n_T$ & Type \\
    \midrule
    California
        & latitude $<36$ $\rightarrow$ latitude $\geq36$
        & 11,814 & 8,826 & spatial \\
    Bike
        & 2011 $\rightarrow$ 2012
        & 8,645 & 8,734 & temporal \\
    Superconduct
        & remaining clusters $\rightarrow$ isolated cluster
        & 16,583 & 4,680 & feature-cluster \\
    PM2.5
        & 8 urban $\rightarrow$ 4 suburban/rural stations
        & 60,000 & 50,000 & spatial \\
    Wine
        & white $\rightarrow$ red wine
        & 4,898 & 1,599 & domain \\
    Parkinsons
        & 28 patients $\rightarrow$ 14 held-out patients
        & 3,932 & 1,943 & domain \\
    KC House
        & latitude $<47.5$ $\rightarrow$ latitude $\geq47.5$
        & 6,607 & 15,006 & spatial \\
    Appliances
        & first two thirds $\rightarrow$ final third
        & 13,156 & 6,579 & temporal \\
    Gas Turbine
        & 2011--2014 $\rightarrow$ 2015
        & 29,349 & 7,384 & temporal \\
    ACS Income
        & California $\rightarrow$ Mississippi and West Virginia
        & 60,000 & 21,292 & domain \\
    Flights
        & months 1--6 $\rightarrow$ months 7--12
        & 60,000 & 50,000 & temporal \\
    Electricity
        & 2012--2013 $\rightarrow$ 2014
        & 17,376 & 8,761 & temporal \\
    Protein
        & remaining clusters $\rightarrow$ isolated cluster
        & 41,457 & 4,273 & feature-cluster \\
    News
        & first two thirds $\rightarrow$ final third
        & 26,429 & 13,215 & temporal \\
    Traffic
        & first two thirds $\rightarrow$ final third
        & 32,136 & 16,068 & temporal \\
    Seoul Bike
        & months 1--8 $\rightarrow$ months 9--12
        & 5,832 & 2,928 & temporal \\
    \bottomrule
\end{tabularx}
\end{table}
\FloatBarrier

\subsection{Target-probe allocation}
\label{app:probe-allocation}

Each dataset provides a labeled target sample of size $n_T$. For each seed, a seeded random permutation is generated. The first 512 indices in the permuted order are assigned to the target probe, and all remaining indices form the test set. The probe budget is fixed across datasets rather than defined as a fraction of $n_T$. The three experimental seeds are $\{0,1,2\}$.

The probe is divided into the three disjoint subsets shown in \cref{tab:probe-allocation}. The diagnostic receives 40\% of the probe. The remaining 307 observations form the method sample, which is divided 70/30 between candidate fitting and gate validation. Rounding produces the allocation below.

\begin{table}[t]
    \centering
    \caption{Allocation of the fixed 512-point labeled target probe.}
    \small
    \begin{tabularx}{0.92\linewidth}{@{}l r X@{}}
    \toprule
    Subset & Size & Permitted use \\
    \midrule
    $\mathcal P_D$
        & 205
        & estimate $\Dcf$ by cross-fitting static and regionwise convex
          fits, then refit both C1 combinations on all 205 observations for
          test evaluation \\
    $\mathcal P_C$
        & 215
        & fit the static convex floor and the six selector candidates \\
    $\mathcal P_G$
        & 92
        & compare the fitted candidates with the floor and make the
          deployment decision \\
    \midrule
    Total
        & 512
        & labeled target-probe budget \\
    \bottomrule
    \end{tabularx}
    \label{tab:probe-allocation}
\end{table}

Within $\mathcal P_D$, $\Dcf$ uses three repetitions of stratified five-fold cross-fitting. In each fold, both the static convex and regionwise convex combiners are fitted on the training folds and evaluated on the held-out fold. Their paired out-of-fold loss
difference forms the diagnostic estimate. For C1 test evaluation, both combiners are then refitted on all 205 observations in $\mathcal P_D$ and evaluated on the disjoint test set.

The $J=8$ partition is learned from the covariates of all 512 probe observations. Partition fitting uses no target labels. The labels in $\mathcal P_D$, $\mathcal P_C$, and $\mathcal P_G$ remain restricted to the uses in \cref{tab:probe-allocation}. In particular, candidate fitting does not use gate labels.

The test set contains the remaining $n_T-512$ target observations. No test covariates or labels enter partition fitting, diagnostic estimation, candidate fitting, or deployment selection. Test labels are available only to the evaluator after the method has made its decision. The smallest test set contains 1,087 observations, so every dataset satisfies the pre-specified minimum of 1,000 test observations.
\FloatBarrier
\subsection{Model-pool construction}
\label{app:model-pool}

The primary model pool contains $K=5$ regression models implemented with scikit-learn \citep{pedregosa2011scikit}. For each dataset, the source sample is divided 80/20 into source-training and source-validation sets. Temporal datasets use a chronological split; the remaining datasets use a seeded random split.

Continuous inputs are standardized using the source-training mean and standard deviation. Deterministic ordinal encodings are applied to categorical inputs before standardization. The target transformation listed in \cref{tab:dataset-sources} is applied before fitting, after which the outcome is standardized as 
\begin{equation}
    y^{\mathrm{std}}=\frac{y-\overline y_S}{s_S},
    \label{eq:target-standardization}
\end{equation}

where $\overline y_S$ and $s_S$ are computed from the source-training outcomes. All reported MSE values use these standardized outcome units. \Cref{tab:primary-pool} gives the primary heterogeneous pool. Parameters not listed in the table use the scikit-learn defaults fixed by the released software environment.

\begin{table}[t]
    \centering
    \caption{Models in the primary heterogeneous regression pool.}
    \small
    \begin{tabularx}{\linewidth}{@{}
    >{\raggedright\arraybackslash}p{0.22\linewidth}
    >{\raggedright\arraybackslash}p{0.33\linewidth}
    >{\raggedright\arraybackslash}X@{}}
    \toprule
    Member & Estimator & Main settings \\
    \midrule
    Gradient boosting
        & \texttt{HistGradientBoostingRegressor}
        & library defaults; seeded \\
    Random forest
        & \texttt{RandomForestRegressor}
        & 200 trees; minimum leaf size 2 \\
    Ridge regression
        & \texttt{Ridge}
        & penalty $\alpha=1$ \\
    Multilayer perceptron
        & \texttt{MLPRegressor}
        & hidden layers $(128,64)$; early stopping; at most 400 iterations \\
    Nearest neighbors
        & \texttt{KNeighborsRegressor}
        & 15 neighbors; distance weighting \\
    \bottomrule
    \end{tabularx}
    \label{tab:primary-pool}
\end{table}
Each mean model is fitted on the source-training set. The source-validation set is used to identify the best source model and to fit one residual-variance head for each pool member. For model $k$, let \[r_{ik}=y_i^{\mathrm{std}}-\widehat\mu_k(x_i)\] denote its source-validation residual. A depth-three histogram gradient-boosting regressor predicts $\log(r_{ik}^2+10^{-8})$ from $x_i$. Its output defines
\begin{equation}
\widehat\sigma_k^2(x)=\exp\left\{\max\!\left(\widehat h_k(x),-10\right)\right\},
\label{eq:variance-head}
\end{equation} 
where $\widehat h_k$ is the fitted log-variance head. These estimates support the uncertainty-weighted baselines; they do not alter the mean predictions. 

The primary uniform baseline averages the five members of the heterogeneous pool. We also construct a homogeneous comparison pool containing five multilayer perceptrons with the same $(128,64)$ architecture and different initialization seeds. This separates the behavior of a heterogeneous model pool from that of a same-architecture deep ensemble. 

After source training and validation, all pool members and variance heads are frozen. Target observations can change ensemble weights, but they cannot update the parameters of a source model. 
\FloatBarrier
\subsection{Preregistration design}
\label{app:preregistration}

We used three within-project preregistration stages. Each record fixed the datasets, hypotheses, evaluation rules, and reporting requirements before the corresponding outcomes were evaluated. The development suite was used to construct the diagnostic. Held-out batches H1 and H2 completed the frozen 12-pair C1 analysis together with the five development pairs, while P3 was reserved for prospective validation of the final selector. Table~\ref{tab:preregistration-stages} summarizes these stages.

\begin{table}[!h]
    \centering
    \small
    \caption{Preregistration stages and their roles in the analysis. H1 and H2 belong to the frozen C1 suite. P3 evaluates the final selector, its diagnostic results enter only the 16-pair sensitivity analysis.}
    \begin{tabularx}{\linewidth}{@{}p{0.08\linewidth}p{0.13\linewidth}p{0.30\linewidth}X@{}}
    \toprule
    Stage & Frozen & Dataset/shift pairs & Role \\
    \midrule
    H1 &
    Aug.\ 3, 2026 &
    Parkinsons, KC House, Appliances, Gas Turbine, ACS Income &
    First held-out test of the diagnostic and its sign-based decision
    rules. \\
    
    H2 &
    Aug.\ 4, 2026 &
    Flights, Electricity &
    Second held-out test under the unchanged diagnostic protocol. \\
    
    P3 &
    Aug.\ 4, 2026 &
    Protein, News, Traffic, Seoul Bike &
    Prospective evaluation of the final Probe-Validated Ensemble Selector.
    C1 quantities are reported only as a sensitivity analysis. \\
    \bottomrule
    \end{tabularx}
    \label{tab:preregistration-stages}
\end{table}
\FloatBarrier
\paragraph{Frozen diagnostic rules.}
For H1 and H2, the target-probe budget, split, partition, loss, model pool, and random seeds were fixed before evaluation. In particular, the protocol used $n=512$, $J=8$, standardized squared error, five-fold cross-fitting repeated three times, and seeds $\{0,1,2\}$. Two diagnostic decision rules were recorded:
\begin{equation}
    \mathcal{R}_1 =
    \begin{cases}
    \mathrm{regional}, & \Dcf > 0,\\
    \mathrm{static},   & \Dcf \leq 0,
    \end{cases}
    \qquad
    \mathcal{R}_2 =
    \begin{cases}
    \mathrm{regional},
    & \operatorname{LCB}_{95}(\Dcf) > 0,\\
    \mathrm{static},
    & \operatorname{LCB}_{95}(\Dcf) \leq 0.
    \end{cases}
    \label{eq:preregistered-rules}
    \end{equation}
For $\mathcal{R}_2$, the lower confidence bound was computed from the 15 paired fold-level loss differences. A realized test gain satisfying 
\[
\left|\widehat G_{\mathrm{test}}\right | <0.002
\]
was classified as indeterminate and excluded from sign hit or miss counts, but retained in the complete results.
Before H1, we recorded weak positive expectations for Parkinsons, KC House, and ACS Income, and nonpositive expectations for Appliances and Gas Turbine. Before H2, both Flights and Electricity were assigned nonpositive expectations. Gas Turbine and ACS Income produced the two reversals reported in the main text.
\paragraph{Prospective selector criteria.} P3 was preregistered after the selector and its six-candidate set had been frozen. The four P3 datasets were not used to design the selector. The preregistration specified three acceptance criteria:
\begin{enumerate}
    \item The worst deployed result could not exceed the static-convex floor by more than $2\%$ in relative loss or $0.002$ in absolute standardized MSE.
    \item Every deployment had to achieve test gain greater than $-0.002$, and the mean gain across deployments had to be positive.
    \item All seeds, gate decisions, candidate losses, fallbacks, and failed runs had to be retained.
\end{enumerate}
The recorded expectations were positive or neutral for Protein, nonpositive for News, uncertain for Traffic, and nonpositive for Seoul Bike. P3 satisfied all three acceptance criteria, as reported in \cref{sec:selector}.
\paragraph{Status and amendments.} These records are "pre-registered within the project" in the sense described in \cref{sec:conclusion}: the research hypotheses, protocols and acceptance criteria were frozen before the corresponding outcomes were observed. Before P3 evaluated any datasets, a revision was made to rename "static unconstrained stacking" to "static affine stacking" and to clarify that the previous analysis of 36 runs on selectors was retrospective. Dated preregistration records and revision logs are included in the artifact.

\FloatBarrier
\section{Synthetic experiment details}
\label{app:synthetic}
\subsection{Generator construction}
\label{app:synthetic-generator}
The controlled experiments use the Bike Sharing temporal shift as their base. For each pool seed, we retain the target outcomes, covariates, and predictions of the source-trained model pool. The generator modifies the predictions while leaving the target outcomes unchanged. 
Generator regions are obtained by applying $k$-means with $J_\star=8$ and seed 777 to the standardized target covariates. These regions define the data-generating structure and are not provided to the ensemble method. Let $r_i\in\{1,\ldots,J_\star\}$ be the generator region of observation $i$, and let \[
\overline\mu_i=\frac{1}{K}\sum_{k=1}^{K}\mu_{ik}.
\]
Base-pool homogenization is applied through 
\begin{equation}
    \mu_{ik}^{(\rho)}=(1-\rho)\mu_{ik}+\rho\overline\mu_i.
\label{eq:synthetic-homogenization}
\end{equation}
Thus, $\rho=0$ preserves the original pool and $\rho=1$ makes all model predictions identical before the synthetic damage is added. The heterogeneity and sparing functions used in 
\cref{eq:synthetic-shift} are 
\begin{align}
    \eta_r(h)&=(1-h)+h(-1)^{r-1},
    \label{eq:synthetic-direction}\\
    S_{rk}(h)&=(1-h)\mathbbm{1}\{k=1\}+h\mathbbm{1}
    \left\{k=1+\bigl((r-1)\bmod K\bigr)
    \right\}
    \label{eq:synthetic=sparing}
\end{align}
The shifted prediction is therefore
\begin{equation}
    \widetilde\mu_{ik}=\mu_{ik}^{(\rho)}+d\,\eta_{r_i}(h)
    \left[1-\kappa S_{r_i k}(h)\right].
\label{eq:synthetic-generator-full}
\end{equation}

At $h=0$, the damage direction is constant and model 1 is spared in every region. At $h=1$, the direction alternates and the spared model rotates across regions. The parameter $\kappa$ controls local sparing: when $\kappa=0$, every model receives the same regional damage; when $\kappa=1$, the selected model is fully spared. The parameter $d$ controls damage magnitude. 

The method does not observe $r_i$. For each run, it learns a separate $J=8$ partition from the available probe covariates. This distinction allows partition identification and approximation error to enter the realized results. 

\subsection{Parameter grid}
\label{app:synthetic-grid}
The four experimental sweeps are listed in \cref{tab:synthetic-grid}. The factorial surface varies $h$ and $\kappa$ over the same five-point grid. Each dose-response experiment varies one quantity while fixing the remaining controls.

\begin{table}[t]
    \small
    \centering
    \caption{Parameter grid for the controlled experiments.}
    \begin{tabularx}{\linewidth}{@{}p{0.22\linewidth}p{0.34\linewidth}X@{}}
    \toprule
    Experiment & Varied values & Fixed values \\
    \midrule
    Factorial surface
    &
    $h,\kappa\in\{0,0.25,0.5,0.75,1\}$&$d=0.5$, $\rho=0$, $n=512$\\
    Shift severity
    &
    $d\in\{0.125,0.25,0.5,0.75,1\}$&$h=\kappa=1$, $\rho=0$, $n=512$\\
    Probe budget
    &
    $n\in\{64,128,256,512,1024\}$&
    $h=\kappa=1$, $d=0.5$, $\rho=0$\\
    Pool homogenization
    &
    $\rho\in\{0,0.25,0.5,0.75,1\}$&$h=\kappa=1$, $d=0.5$, $n=512$\\
    \bottomrule
    \end{tabularx}
    \label{tab:synthetic-grid}
\end{table}
The primary design uses 10 model-pool seeds and three target-split seeds per pool. The three splits are averaged within each pool, so the 10 independently trained pools are the units of inference. Parameter combinations shared by more than one sweep are counted once, leaving 1110 distinct run cells. For the budget experiment, each pool and split seed produces one permutation of the target observations. The probes are nested prefixes, \[
\mathcal P_{64}
\subset
\mathcal P_{128}
\subset
\mathcal P_{256}
\subset
\mathcal P_{512}
\subset
\mathcal P_{1024},
\]
and the test set is fixed as the observations after position 1024 in the permutation. Changes along the budget curve therefore arise from the amount of probe information rather than changes in test composition. Each probe follows the diagnostic, combiner, and gate proportions defined in \cref{app:probe-allocation}.

The realized-gain curves use the three-candidate selector frozen for the synthetic experiment. Its candidates are the Shrunken Regional Convex Combiner, a smooth covariate gate, and probe-fit inverse-variance weighting. All candidates are fitted on the combiner split and validated on the gate split against a static-convex floor. The final six-candidate deployment procedure is described in \cref{sec:selector,app:selector-details}.
\subsection{Pool-level inference}
\label{app:synthetic-inference}

For each pool, we first average the three target splits within every $(h,\kappa)$ cell. We then fit 
\begin{equation}
    G_{p}(h,\kappa)=\beta_{0p}+\beta_{hp}h+\beta_{\kappa p}\kappa+\beta_{h\kappa,p}h\kappa+\varepsilon_{p}(h,\kappa)
\label{eq:synthetic-interaction}
\end{equation}
across the 25 factorial cells. The reported interaction is the mean of the 10 independently estimated $\beta_{h\kappa,p}$ values. Confidence intervals and tests use a one-sample $t$ distribution with nine degrees of freedom. The same procedure is applied to realized gain, $\Dcf$, and the oracle gain.
For realized gain, this gives
\[
\beta_{h\kappa}=0.0830,
\qquad
95\%~\mathrm{CI}=[0.0760,0.0900],
\qquad
t_9=26.8.
\]
The interaction estimates are $0.0842$ for the oracle gain and $0.0837$ for $\Dcf$. The corresponding 95\% intervals are $[0.0770,0.0914]$ and $[0.0698,0.0976]$.
An earlier sensitivity run used three model pools and 10 target splits per pool. Its realized gain at $(h,\kappa)=(1,1)$ was $0.0864$, compared with $0.0910$ in the primary 10-pool design.
\subsection{Probe-budget decomposition}
\label{app:synthetic-budget}
The budget experiment distinguishes three quantities:
\begin{enumerate}
    \item The structural ceiling uses the generator regions and fits convex weights directly on the fixed test outcomes. It is unavailable to a deployed method and is constant in $n$ by construction.
    \item The partition-conditional oracle uses the partition learned from the probe but fits its convex weights on the fixed test outcomes. Its gap from the structural ceiling contains partition estimation error and mismatch between the $k$-means partition class and the generator regions.
    \item Realized gain uses only the probe to learn the partition, estimate candidate parameters, and make the gate decision.
\end{enumerate}
The structural ceiling is $0.1225\pm0.0067$, using a 95\% $t_9$ interval. The partition-conditional oracle increases from $0.099$ at $n=64$ to $0.112$ at $n=1024$. Realized gain increases from $0.005$ to $0.098$. The first gap measures partition identification and approximation cost, the second adds weight-estimation and validation cost. 
In the homogenization experiment, moving from an identical pool ($\rho=1$) to the original diverse pool ($\rho=0$) reduces the static-oracle loss by $0.0576$ and the regional-oracle loss by $0.0221$. Generic diversity therefore benefits the static oracle about 2.6 times as much in this experiment. The regional advantage is instead generated by the conditional specialization controlled by $h$ and $\kappa$.

An initial generator version alternated the damage direction even when $h=0$, leaving regional structure active in a cell intended to contain no heterogeneity. In the frozen generator, \cref{eq:synthetic-direction} links direction heterogeneity to $h$. All primary results in \cref{fig:c3} use this corrected definition.

\FloatBarrier

\section{Selector and baseline implementations}
\label{app:selector-details}
This appendix documents the specific implementations of the various combiners used in the paper. All fitting operations are performed on standardized target values (using the mean and scaling parameters from the source domain training set) and are restricted to data from the "method-probe split"; the diagnostic split $\mathcal P_D$ remains untouched by any deployed combiners, while the test split is strictly excluded from any fitting or gating steps.

\subsection{The static convex floor}
The “floor” $O_{\mathrm{static}}$ represents the best fixed combination result achievable via a convex combination of $K$ prediction sources for a given labeled dataset; it is obtained by solving $\min_{w \in \Delta^{K-1}} \tfrac{1}{n}\lVert M w - y\rVert^2$, where $M \in \mathbb{R}^{n \times K}$ contains the mean predictions from each source. We employ a “support enumeration” method to solve this problem exactly: for each of the $2^K - 1$ non-empty supports, we solve the corresponding equality-constrained KKT system, discard solutions containing negative components, and retain feasible minimizers. For $K=5$, this entails solving 31 small systems of linear equations; notably, this is an exact solution method rather than an iterative one. We verified the accuracy of this exact solver by comparing it against exponentiated gradient descent on the simplex (3,000 iterations, learning rate 0.5), finding agreement to four decimal places; this solver is widely applied in scenarios where $K \le 10$.

The same solver is also used to define the “region-level convex combiner”: the simplex optimization problem is solved independently for each region, whereas the global solution is retained for regions containing fewer than $m_{\min} = 5$ labeled probe points. Since refining a single global convex combination into independent regional convex combinations can only reduce in-sample loss, the resulting “headroom” is necessarily non-negative by construction.
\subsection{Candidate menu}
This selector is agnostic to the specific candidate schemes: it validates any input menu against a "floor." This fixed menu comprises six members, all fitted on the "method-train probe" (approximately 215 data points) using hyperparameters that were set prior to downloading the held-out batches.
\begin{enumerate}
    \item Regional simplex solutions shrink toward the global solution: $w_r \leftarrow \lambda_r w_r + (1-\lambda_r) w_{\mathrm{glob}}$, where $\lambda_r = n_r/(n_r + n_0)$ and $n_0 = 10$; for regions with $n_r < 5$, $w_{\mathrm{glob}}$ is used directly.
    \item We employ covariate-dependent stacking \citep{wakayama2025ensemble} with affine, non-simplex weight functions of the form $w_k(x) = \mu_k + E(x)^\top \gamma_k$, constructed using $M$-dimensional Gaussian RBF basis functions; here, $M = \min(16, \max(2, \lfloor n/10 \rfloor))$, with basis function centers determined by the $k$-means algorithm and bandwidths set via the median pairwise distance heuristic. The penalty terms for each model are estimated using the EM algorithm (assuming $\gamma_k \sim \mathcal{N}(0, \tau_k^2 I)$ and $\lambda_k = \sigma^2/\tau_k^2$) over 100 iterations, with an analytical solution available for the M-step. Verification of the method's fidelity is detailed in Appendix \S\ref{app:cdst-checks}.
    \item A single hidden layer with 32 $\tanh$ units feeds into a softmax output layer representing $K$ experts. Training employs full-batch gradient descent with a momentum of 0.9, a learning rate of 0.03, and an $\ell_2$ penalty of $10^{-4}$ over 1,500 iterations. The experts are members of a frozen pool; as access protocols prohibit retraining them, this constitutes a Mixture of Experts (MoE) variant where only the gating network is trained, rather than an end-to-end MoE.
    \item The variance prediction head for each model is refitted on a "method probe". That is, depth 2 histogram gradient boosting based on $\log((y - \mu_k)^2 + 10^{-8})$, using weights proportional to $1/\hat\sigma_k^2(x)$, normalized to simplex; the lower bound on log variance predictions is set to $-10$.
    \item Softmax weights linear in the covariates, $w(x) \propto \exp(\tilde{x}^\top W)$ with an intercept column, fitted by 800 gradient steps on the stacked squared loss, learning rate $0.05$, $\ell_2$ penalty $10^{-3}$.
    \item Unconstrained least squares with an intercept term—specifically, $\hat y = b + \sum_k a_k \mu_k$ (i.e., classical stacked regression without simplex constraints). This method is structurally independent of the inputs; it is introduced precisely to distinguish between "adapting to $x$" and "correcting the global level and scale."
\end{enumerate}
Finally, the affine candidate method is precisely what creates a non-nested relationship between this set of methods and the baseline (floor): during development runs, the row sums of the fitted CDST-RBF weight vectors ranged from $0.48$ to $1.68$, with individual components dropping as low as $-3.3$—indicating that the method performs contraction and cancellation operations rather than simple mixing. Consequently, the paper emphasizes that the critical constraint distinguishing these candidate methods is the simplex constraint, rather than whether or not they rely on input data.

\subsection{The validation gate}
Let $\ell_{\mathrm{sta}}(i)$ and $\ell_c(i)$ denote the "floor" (baseline lower bound) and the squared error of candidate model $c$ on the "gate split" (comprising approximately 92 data points, disjoint from all fitting sets), respectively, and let $d_c(i) = \ell_{\mathrm{sta}}(i) - \ell_c(i)$. Among candidate models satisfying $\overline{d_c} > 0$, the selector chooses $c^\star = \arg\max_c \overline{d_c}$ and deploys the model only if the following condition holds:
\begin{equation}
\mathrm{LCB}(c^\star) \;=\; \overline{d_{c^\star}}
\;-\; t\,\frac{\mathrm{sd}(d_{c^\star})}{\sqrt{n_g}} \;>\; 0 ,
\end{equation}
otherwise, the static convex combination floor is deployed. This threshold is based on the one-sided $95\%$ Student's t-quantile with $\mathrm{df} \approx 91$ degrees of freedom, adjusted via Bonferroni correction for the size of the candidate model set: $t = 2.42$ for a set of six candidate models, and $t = 2.13$ for the set of three candidate models used in the synthetic data study (see Appendix \ref{app:synthetic}). The floor itself is deployed unconditionally: it consumes 4 degrees of freedom across approximately 215 data points and requires no certificate. An earlier development version included a gating mechanism for the floor against the "uniform average" and allowed a fallback to the uniform average; this intermediate gating mechanism resulted in a normalized mean squared error (MSE) loss of 0.03 on the \texttt{pm25\_site} dataset and was therefore removed prior to processing held-out batches.

Note that the selection statistic and the certificate are computed based on the same gate split. The Bonferroni correction controls for selection effects within the candidate set; the nominal level associated with the t-statistic-based threshold is merely an approximation, and the empirical "harmlessness" documented in the paper forms the basis for our claim.

\subsection{The three-candidate selector of the synthetic study}
These synthetic experiments predated the proposal of the "six candidate schemes" and employed the v1-version selectors: specifically, the "Shrunken Regional Convex Combiner" (with parameter $n_0 = 10$), the linear softmax covariate gate, and "probe-fit inverse-variance weighting." All schemes utilized gating against the same static convex floor at $t = 2.13$ (one-sided $95\%/3$ confidence level, degrees of freedom $\mathrm{df} \approx 91$). The candidate scheme labeled "smooth covariate gate" corresponds to the linear softmax gate described in item 5 above; it is not the CDST-RBF estimator, which was only included when the "six candidate schemes" were introduced.

\subsection{Faithfulness of the CDST-RBF implementation}
\label{app:cdst-checks}
Given that the CDST estimator significantly influences our conclusions, we validated it before utilizing its numerical results. When formulated as a linear mixed model $y = F\mu + Z\gamma + \varepsilon$ (where $F = M$ and $Z[i,(k,m)] = E_m(x_i)\,\mu_k(x_i)$), the penalty term in the paper's objective function, $\sum_k \lambda_k \gamma_k^\top \gamma_k$, corresponds exactly to the random-effects prior with $\lambda_k = \sigma^2/\tau_k^2$; consequently, the EM algorithm described in the paper is directly applicable. Specifically, the validation comprises four checks (corresponding file: \texttt{experiments/test\_cdst\_em.py}):
\begin{enumerate}
    \item The observed-data log-likelihood is non-decreasing at all 61 recorded iterations (zero decreases).
    \item Based on data simulated using the model from the paper, the correlation coefficient between the reconstructed weight surface and the original surface is $0.911$. Since both the intercept and the basis coefficients depend on the specific representation and the basis used for fitting differs from the one used to generate the data, the identifiable target is the surface itself rather than these coefficients.
    \item The objective function in the paper employs the base prediction $f_{k,-i}(x_i)$ obtained by refitting the model after excluding observation $i$. Under our experimental setup, the ensemble of base models is trained on "source" data, while the stacking process is performed on a disjoint "target" probe set; consequently, the observations used for stacking are not included in the training sets of any base models, rendering $f_{k,-i}(x_i)$ identical to $f_k(x_i)$ (i.e., $\max_i |f_k - f_{k,-i}|$ is zero within machine precision). Thus, our objective function is exactly equivalent to the one in the paper, rather than merely an approximation.
    \item Upon convergence of the EM algorithm, the penalized objective function becomes a convex quadratic function with an analytical solution (i.e., a closed-form minimizer); the fixed-point solution of the EM algorithm reaches this minimizer, with a relative error of only $1.3 \times 10^{-6}$. Twenty random perturbation tests conducted on $(\tau^2, \sigma^2)$ showed no increase in the marginal log-likelihood.
\end{enumerate}

There are two documented and validated improvements: the aforementioned "Leave-One-Out" (LOO) procedure and the use of a median heuristic (rather than manual specification) to set the basis function bandwidth.

\section{Reproducibility}
\label{app:reproducibility}
All experiments were conducted on a laptop equipped with a CPU, with fixed dependency versions (\texttt{numpy}~2.5.1, \texttt{scikit-learn}~1.9.0, \texttt{scipy}~1.18.0, \texttt{pandas}~3.0.5, \texttt{folktables}~0.0.12) and random seeds $\{0, 1, 2\}$. The entire experimental pipeline can be re-run end-to-end on a single machine within one hour; the most time-consuming task was the C3 factor grid experiment (approximately 33 minutes).

Each data loader downloads data from the original source and verifies checksums; a single driver script manages the re-execution of all experiments and validates six key metrics against an automated "golden standard" check. Re-runs in a "clean-room" environment reproduced all 30 key metrics from the development phase (5 datasets $\times$ 3 random seeds $\times$ 2 statistical metrics), with deviations within $0.002$.

The 12 UCI datasets are released under the CC~BY~4.0 license (attribution details are provided in Appendix~\ref{app:datasets}); \texttt{kc\_house} (OpenML) and \texttt{nycflights13} are released under the CC0 license. The ACS tasks are built upon the MIT-licensed \texttt{folktables} package \citep{ding2021retiring}; the use of the underlying PUMS data complies with the U.S. Census Bureau's terms of service. The California Housing data (StatLib) lacks an explicit license, so we utilize the data without redistributing it; the experimental artifact does not contain any data files.

\section{Code availability}
The OpenRegShift artifact: pool construction, diagnostic, selector, experiment scripts, preregistration documents and result manifests, will be released publicly upon publication.

\end{document}